\documentclass{article}

\usepackage[preprint]{neurips_2026}

\usepackage{enumitem}
\usepackage[utf8]{inputenc} 
\usepackage[T1]{fontenc}    
\usepackage{hyperref}       
\usepackage{url}            
\usepackage{booktabs}       
\usepackage{amsfonts}       
\usepackage{nicefrac}       
\usepackage{microtype}      
\usepackage{graphicx}      
\usepackage{amsmath}
\usepackage{xcolor}
\usepackage{listings}
\usepackage{longtable}
\usepackage{wrapfig}
\usepackage{multirow}
\usepackage[table]{xcolor}
\usepackage{graphicx}
\usepackage{subcaption}
\usepackage{pifont}
\newcommand{\cmark}{\ding{51}}

\definecolor{headergray}{gray}{0.92}
\definecolor{bestred}{RGB}{255, 220, 220}
\definecolor{secondgreen}{RGB}{220, 245, 220}

\usepackage{amsmath}
\usepackage{amssymb}
\usepackage{mathtools}
\usepackage{amsthm}
\usepackage{tcolorbox}
\usepackage[capitalize,noabbrev]{cleveref}

\theoremstyle{plain}

\theoremstyle{definition}

\theoremstyle{remark}

\title{Do LLMs Truly Generalize in the Molecular Domain? A Perturbation-Based Analysis}

\author{%
  Jiatong Li$^{1,2}$ \quad Weida Wang$^{3,4}$ \quad Changmeng Zheng$^1$ \quad Shufei Zhang$^3$\\
  \textbf{Yatao Bian}$^2$ \quad \textbf{Xiao-yong Wei}$^{1\dagger}$ \quad \textbf{Qing Li}$^1$  \\[0.8ex]
  $^1$The Hong Kong Polytechnic University \quad
  $^2$National University of Singapore \\
  $^3$Shanghai AI Lab \quad
  $^4$Fudan University \\
  $^\dagger$Corresponding author: \texttt{x1wei@polyu.edu.hk}
}

\begin{document}

\maketitle

\begin{abstract}

Large Language Models (LLMs) have recently shown promise in molecular discovery, yet a gap remains between their probabilistic nature over discrete sequential tokens and the rigid topological constraints of chemical space. This raises the question of whether molecular LLMs can generalize beyond the local neighborhoods induced by their sequence-based representations. To systematically investigate this question, we introduce a \textbf{Molecular Perturbation} framework that generates syntax-valid structural variants of training molecules under controlled Graph Edit Distance (GED) to probe the manifold regularity of molecular LLMs. Our analysis shows that even a single edit can cause substantial performance drops on common molecular tasks, revealing a narrow local \emph{trust region} and fragile sensitivity to structural changes. 
Since similar molecules tend to exhibit similar properties, In-Context Tuning (ICT), which anchors predictions on structurally similar molecules, offers a natural way to mitigate such fragility.
Our experiments also examine whether ICT confers robustness under controlled structural perturbations, and the results suggest that it can partially expand the local \emph{trust region} and offer a promising direction for stabilizing molecular LLMs against structural variation.

\end{abstract}

\section{Introduction}

\begin{wrapfigure}{r}{0.35\linewidth}
    \centering
    \vskip -0.25in
    \includegraphics[width=\linewidth]{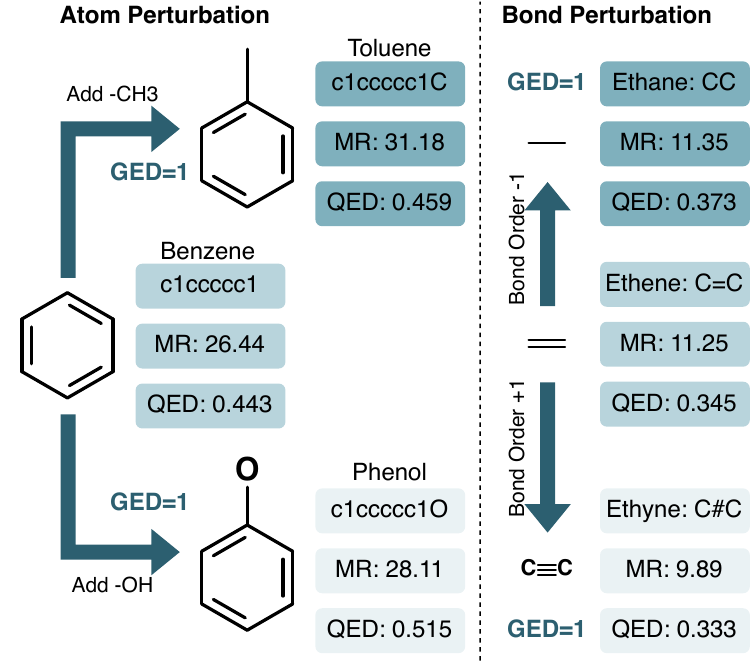}
    \vskip -0.1in
    \caption{Illustration of Molecular Perturbation, including atom perturbations and bond perturbations.}
    \vskip -0.15in
    \label{fig:intro}
\end{wrapfigure}

The integration of Large Language Models (LLMs) into chemical research has enabled models to generate and predict molecular structures with apparent fluency. By linearizing molecular graphs into sequential notations like SMILES \citep{weininger1988smiles} and SELFIES \citep{krenn2020self}, these models leverage the statistical power of token prediction \citep{vaswani2017attention, radford2018improving} to navigate the vastness of chemical space. However, this empirical success masks an ontological disconnect. LLMs are engineered to optimize probabilistic likelihood over discrete sequences, a paradigm well-suited for natural language where semantic meaning is resilient to minor syntactic noise. In stark contrast, chemistry is governed by rigid topological constraints \citep{hubin2003synthesis} and quantum mechanical laws \citep{primas2013chemistry}, where the ``syntax'' is non-negotiable and the ``semantics'' are highly volatile. As shown in Figure~\ref{fig:intro}, a single structural edit does not merely alter the local sequence, but also impacts the molecule's bioactivity, a complex mapping that can be difficult to capture. This raises a critical, often overlooked question: \textit{Do LLMs truly learn the underlying manifold of chemical space via molecular textual representations, or do they merely exploit surface sequence or topological patterns that interpolate within local training clusters?}

To answer this, we must scrutinize the model's behavior not on static benchmarks, but through the lens of \textbf{manifold regularity}. A robust scientific model should exhibit \textit{Lipschitz continuity} relative to the chemical structure, meaning that small perturbations in the input graph should map to predictable, smooth transitions in the latent representation space. We find that current LLMs fail to model this duality even for different textual representations of the same molecule \citep{ganeeva2024chemical}, simply because their training objectives prioritize textual coherence over topological consistency. They lack an intrinsic mechanism to \emph{align} structural-level variation with chemically meaningful similarity, confining generalization to the narrow training distribution. Consequently, when faced with molecules that drift slightly from the training distribution, the performance may not just degrade, but collapse.

To rigorously diagnose this fragility, we dismantle the standard random-split evaluation paradigm, which often suffers from data leakage and fails to test true generalization. Instead, we introduce a systematic evaluation framework centered on \textbf{Molecular Perturbation}. We generate syntax-valid ``structural variants'' under controlled Graph Edit Distance (GED) \citep{gao2010survey}, constructing a spectrum of deformation that probes the model's robustness. By applying granular \textit{Atom Perturbations} (node feature shifts) and \textit{Bond Perturbations} (edge feature shifts), we simulate a walk away from the training manifold. This allows us to dissect the model's failure modes.

Our empirical analysis reveals a sobering reality: for specific tasks, even a single structural edit can cause substantial performance drops, and the degradation remains pronounced as GED increases. This suggests that, while models may preserve some coarse-grained continuity over chemical space, their local smoothness around training molecules is limited (i.e., a narrow \emph{trust region}), consistent with strong reliance on high-density regions of the training distribution. 
A natural question is whether this fragility can be alleviated by exposing the model to chemically related references at inference time, in line with the long-standing molecular similarity principle \citep{bender2004molecular} that similar molecules tend to share similar properties.

This perspective directly connects to In-Context Tuning (ICT) \citep{li2024empowering, chen2025reactgpt}, which fine-tunes molecular LLMs to condition predictions on a small set of structurally similar molecules retrieved from the training set. ICT can be viewed as an operationalization of the molecular similarity principle: rather than treating each molecule as an isolated input requiring global extrapolation from internalized parameters, it reframes prediction as a locally grounded inference anchored on chemically related neighbors. If the fragility we observe stems from over-reliance on high-density regions of the training distribution, then ICT, by explicitly anchoring inference to nearby training examples, should in principle confer robustness as inputs drift away from the training manifold.

We therefore systematically examine ICT through the lens of our Molecular Perturbation framework, tracking how prediction quality evolves as GED increases. Our results suggest that ICT can partially expand the local trust region and alleviate the fragility of direct prediction, with neighbor-conditioned models retaining more performance under controlled structural edits. While ICT does not fully eliminate sensitivity to perturbation, these findings indicate that retrieval-based anchoring offers a promising direction for stabilizing molecular LLMs against structural variation, pointing toward local, similarity-grounded inference as a practical mechanism for bridging the gap between probabilistic sequence modeling and the rigid topology of chemical space.

\section{Related Work}

Early efforts for applying LLMs in molecule discovery predominantly focused on cross-modal retrieval \citep{edwards2021text2mol} and bidirectional translation \citep{edwards2022translation} between natural language and SMILES strings. Models such as MolT5 \citep{edwards2022translation} and KV-PLM \citep{zeng2022deep} established that
treating molecular notations as a specialized language allows LLMs to capture fundamental chemical grammar, effectively bridging the gap between high-level human descriptions and formal chemical notations.   
Subsequent works focused on strengthening the capability of LLMs through domain-specific pre-training and instruction-based alignment. For instance, BioT5 \citep{pei2023biot5} and MolXPT \citep{liu2023molxpt} established strong baselines by pre-training on large-scale molecular corpora and biological literature, often employing ``wrapped text'' paradigms where molecules in scientific papers were replaced with their corresponding sequences to capture deep contextual associations. At the same time, the Mol-Instructions \citep{fangmol} dataset provides a large-scale biomolecular instruction set to refine model proficiency across tasks like retrosynthesis and property prediction. Furthermore, there are also sophisticated benchmarks to probe creative and logical capabilities, such as TOMG-Bench \citep{li2024speak}, ChemLLMBench \citep{guo2023can}, and ChemCoTBench \citep{li2025beyond}.

\section{Molecular Perturbation}
\label{sec:perturbation}
To systematically assess the manifold regularity of molecular LLMs, we propose a Molecular Perturbation strategy to perform minor syntax-valid structural edits on molecular graphs. This strategy allows us to simulate a spectrum of \textit{out-of-distribution} by quantifying the deviation of a perturbed molecule from its original structure via Graph Edit Distance (GED).

\subsection{Perturbation Strategy}
As shown in Figure \ref{fig:intro}, our strategy mainly operates at two fundamental levels of molecular structure (i.e., \textit{Atom Perturbation} and \textit{Bond Perturbation}), which are designed to mimic common chemical modifications that may occur in drug design, such as functionalization and saturation changes.

\noindent\textbf{Atom Perturbation.}
For atom-level perturbations, we perform \textbf{random heteroatom substitution}. Specifically, we randomly select non-carbon atoms in the molecule and replace them with an alternative atom sampled from the set $\{B, N, O, F, P, S, Cl\}$, subject to a simple valence-preserving constraint. 
During substitution, we require that the total valence of the original atom does not exceed the default valence of the newly assigned atom, to avoid invalid configurations. 
This operation introduces minimal yet semantically meaningful structural perturbations while preserving the molecular topology, enabling a controlled evaluation of the model’s local generalization behavior.

\noindent\textbf{Bond Perturbation.}
For bond-level modifications, we perform \textbf{random bond order perturbations} under implicit valence constraints. 
Specifically, hydrogen atoms are first removed to avoid selecting bonds involving hydrogens, and a subset of existing bonds is randomly sampled for modification. 
For each selected bond, we attempt to adjust its bond order while simultaneously updating the associated hydrogen counts to preserve valid valence configurations. 
If the resulting molecular structure violates valence constraints or cannot be sanitized, the modification is discarded and the bond is excluded from further consideration. 
This process continues until a predefined number of successful bond modifications is reached or no valid candidate bonds remain.

\subsubsection{Syntax Validity Check}

A central challenge in automated molecular perturbation lies in maintaining syntax validity. Besides common RDKit sanitization, in our framework, syntax validity is also enforced through an \textbf{incremental rejection mechanism} integrated directly into the perturbation process, rather than via post-hoc filtering:

\noindent For \textbf{atom-level} edits, candidate replacements are accepted only if the total valence of the original atom does not exceed the default valence of the newly assigned element, thereby ensuring basic valence consistency.

\noindent For \textbf{bond-level} modifications, hydrogen atoms are temporarily removed and bond order changes are attempted only when the resulting structure can be successfully sanitized; otherwise, the modification is discarded and excluded from further consideration.

Once a perturbed molecule is obtained, we query its corresponding IUPAC name from database and retain only those that yield a valid result. In contrast to molecule captions, which may admit multiple valid textual descriptions for a single structure, IUPAC nomenclature provides a standardized and unambiguous representation. This ensures that our evaluation focuses strictly on the model’s ability to interpret the chemical semantics under controlled structural perturbations.

\subsection{Graph Edit Distance}

To quantify the magnitude of structural changes induced by our perturbation strategy, we adopt \textbf{Graph Edit Distance (GED)} to control the molecular dissimilarity of variants. 
GED evaluates the similarity between two graphs by computing the minimum-cost sequence of edit operations required to transform one graph into another.

Formally, given an original molecular graph $G_1 = (V_1, E_1)$ and a perturbed molecular graph $G_2 = (V_2, E_2)$, the GED is defined as:
\begin{equation}
GED(G_1, G_2) = \min_{(e_1, \dots, e_k) \in \mathcal{P}(G_1, G_2)} \sum_{i=1}^{k} c(e_i),
\end{equation}
where $\mathcal{P}(G_1, G_2)$ denotes the set of all valid edit paths transforming $G_1$ into $G_2$, and $c(e_i)$ is the cost associated with each node or edge edit operation.

In our framework, we define GED as the cumulative number of \textbf{successful atomic and bond-level edit operations} applied during perturbation stage. 
Each accepted atom substitution or bond type modification contributes a unit cost to the total GED. 
To probe the generalization spectrum, we consider five discrete perturbation levels with $GED \in \{1,2,3,4,5\}$, corresponding to increasingly larger but controlled structural deviations from the original molecule, which ensures a fine-grained analysis of model robustness, allowing us to trace performance degradation from molecular structures with low GED to progressively more altered yet syntax valid derivatives (high GED).

\begin{figure}[t]
    \centering
    \vskip -0.4in
    \begin{subfigure}[b]{0.49\linewidth}
        \centering
        \includegraphics[width=\linewidth]{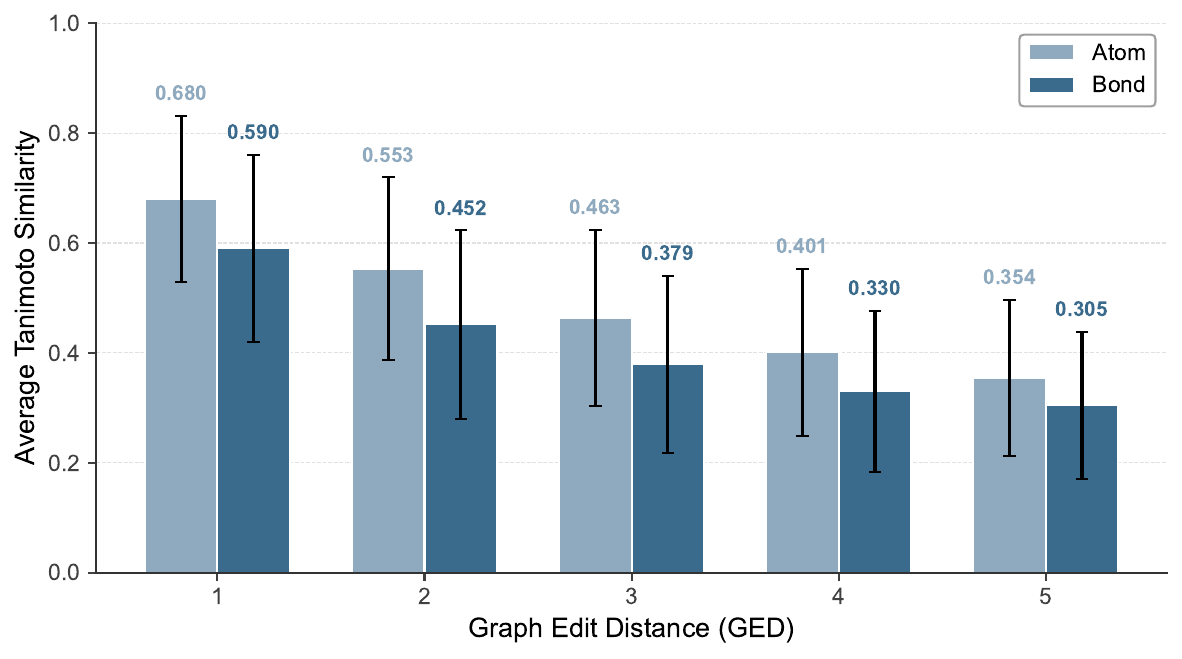}
        \caption{Average Molecular Similarity}
        \label{fig:sim}
    \end{subfigure}
    \hfill
    \begin{subfigure}[b]{0.49\linewidth}
        \centering
        \includegraphics[width=\linewidth]{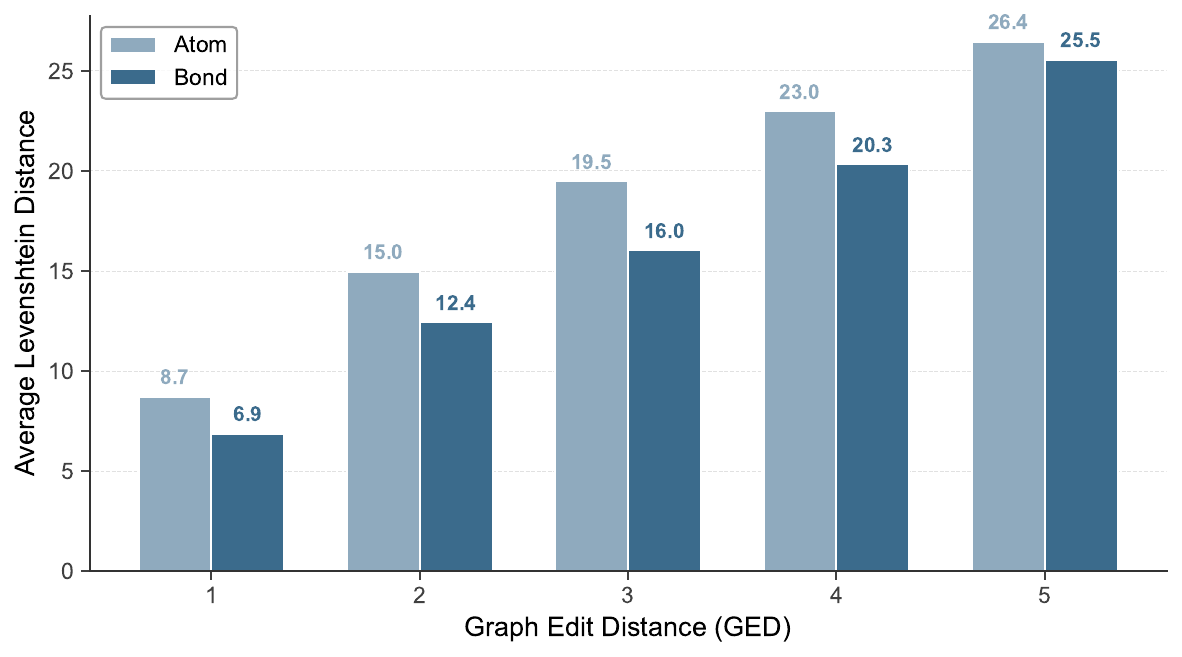}
        \caption{Average Levenshtein distance}
        \label{fig:lev}
    \end{subfigure}
    \vskip -0.05in
    \caption{Comparison of atom-only and bond-only perturbations across five graph edit distance (GED) levels, measured by (a) molecular similarity and (b) Levenshtein distance on SMILES strings.}
    \vskip -0.2in
    \label{fig:sim_lev}
\end{figure}

\subsection{Perturbation Effectiveness}
\label{subsec:similarity}

To validate GED as an effective notion of perturbation strength, we examine whether it correlates with both (i) \emph{chemical} similarity (Tanimoto similarity) and (ii) \emph{sequence-level} difference (Levenshtein distance) between the corresponding molecular strings.
As shown in Figure~\ref{fig:sim}, GED=1 perturbations maintain high structural fidelity (average similarity $>$ 0.6), and chemical similarity decreases monotonically as GED increases.
Consistently, Figure~\ref{fig:lev} shows that the average Levenshtein distance increases with GED, indicating that larger topological edits also induce larger sequence-level changes.
Together, these trends confirm that increasing GED simultaneously moves molecules farther in chemical space and farther in token space, supporting GED as a controlled axis for generalization analysis.

A comparative analysis reveals an intriguing \emph{mismatch} between chemical-space and sequence-space distances. Chemically, bond-level modifications exert a more profound impact on molecules than atom-level perturbations: at GED=1, bond perturbations result in a markedly lower Tanimoto similarity compared to atom substitutions (0.680 vs. 0.590). However, in sequence space, we observe the opposite trend in Figure~\ref{fig:lev}: bond perturbations typically induce \emph{smaller} Levenshtein distance than atom substitutions (8.7 vs. 6.9).

This divergence highlights that ``closeness'' under molecular strings is not a reliable proxy for topological/chemical closeness. In particular, changing a bond order can substantially alter molecular connectivity while minimally editing the string, whereas an atom substitution may require larger textual changes while leaving the scaffold largely intact. 

\section{Manifold Regularity Analysis}
\label{subsec:manifold}

\begin{figure*}[t]
    \centering
    \vskip -0.4in
    \includegraphics[width=\linewidth]{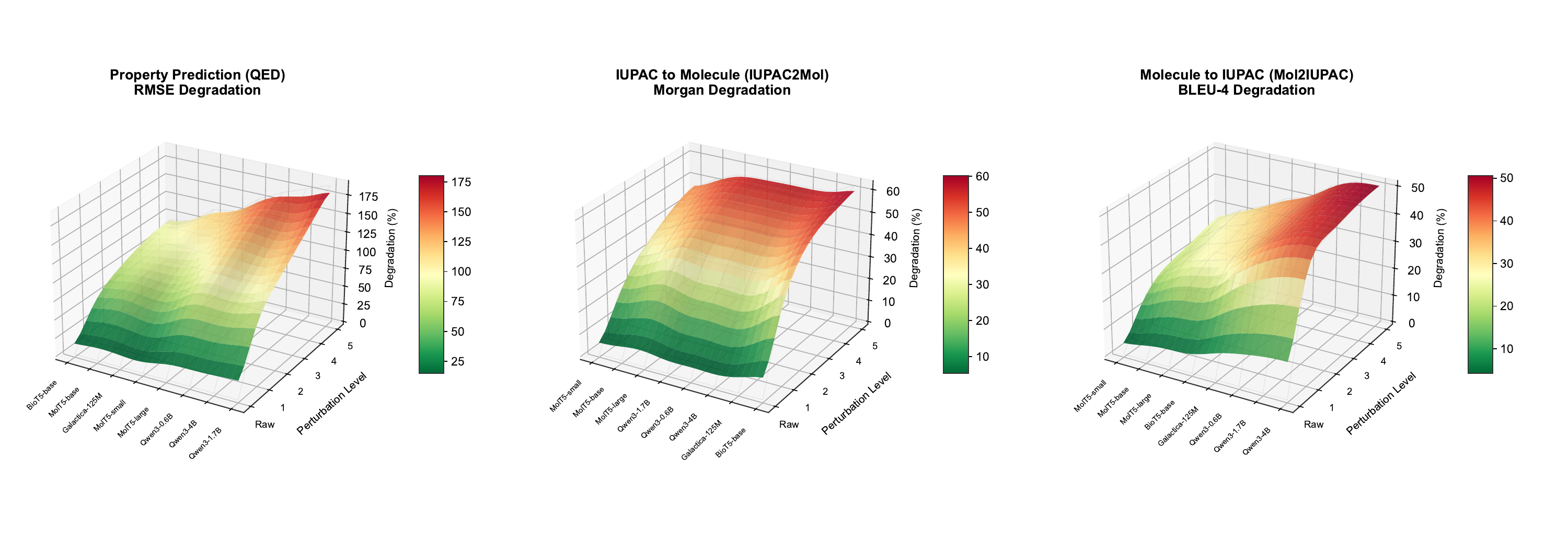} 
    \vskip -0.4in
    \caption{Performance degradation landscapes (smoothed) with increasing perturbation level. Each surface corresponds to one model and reports the relative drop (\%) to \emph{raw\_train} in the task metric (RMSE, Morgan, or BLEU-4); higher values indicate poorer robustness.}
    \vskip -0.1in
    \label{fig:performance_ged}
\end{figure*}

\begin{figure*}[t]
    \centering
    \includegraphics[width=1.0\linewidth]{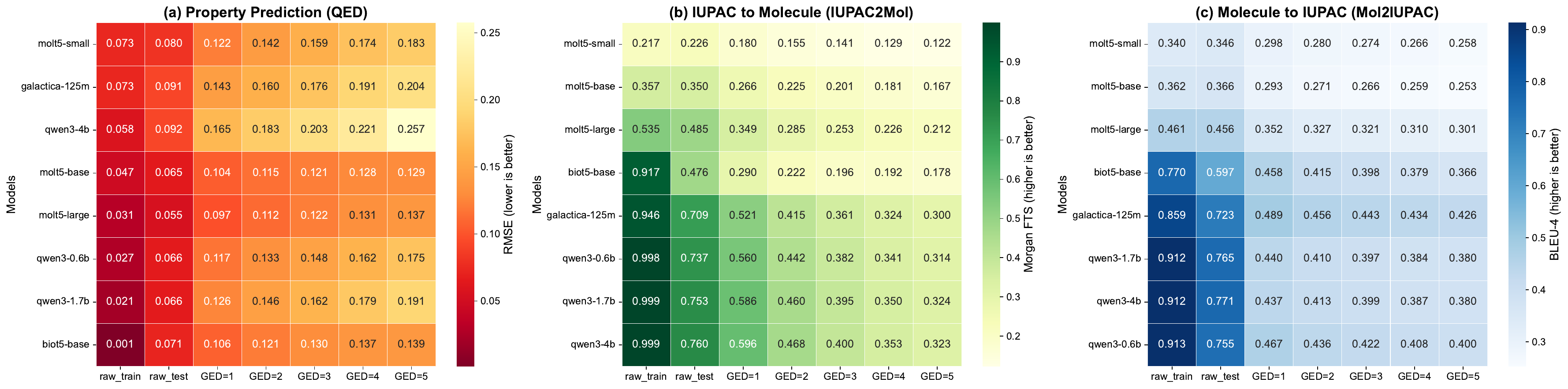}
    \caption{Absolute task performance on the original train/test splits and perturbed test sets (GED from 1 to 5). (a) QED prediction (RMSE, lower is better). (b) IUPAC2Mol (Morgan FTS, higher is better). (c) Mol2IUPAC (BLEU-4, higher is better).}
    \vskip -0.2in
    \label{fig:performance_heat}
\end{figure*}

In this part, we analyze how molecular LLMs generalize under controlled structural perturbations. We first report the experimental setup and demonstrate the overall performance degradation corresponding to increasing GED level (Figure~\ref{fig:performance_ged}) and then provide a comprehensive view of absolute performance across the original train/test and perturbed sets (Figure~\ref{fig:performance_heat}). Finally, we break down robustness by perturbation type (atom vs. bond), model architectures and scales.

\subsection{Experimental Setup}
\label{subsec:exp_setup}

We evaluate manifold regularity by measuring how model performance changes as GED increases ($GED=1\rightarrow5$) across various tasks.

\noindent\textbf{Model Families.}
Generally, we include two model architectures, including decoder-only models and encoder-decoder models. Both general-purpose and domain-adapted models are examined:
\begin{itemize}[leftmargin=*, itemsep=2pt, topsep=2pt, parsep=0pt]
    \item \textbf{Decoder-only:} We include \textbf{Galactica-125M} \citep{taylor2022galactica}, a model pre-trained on large-scale scientific corpora, and the general models like \textbf{Qwen3} series: \textit{0.6B, 1.7B, and 4B} parameters \citep{yang2025qwen3}.
    \item \textbf{Encoder-decoder:} We evaluate the \textbf{MolT5} family: \textit{small, base, and large} \citep{edwards2022translation}. Additionally, we include \textbf{BioT5-base} \citep{pei2023biot5}; notably, BioT5 utilizes the SELFIES representation of molecular structures.
\end{itemize}

\noindent\textbf{Tasks and Metrics.}
The models are examined across three representative chemical downstream tasks, each targeting a different facet of molecular understanding and generation:
\begin{itemize}[leftmargin=*, itemsep=2pt, topsep=2pt, parsep=0pt]
    \item \textbf{QED (drug-likeness prediction):} This task requires the model to predict the Quantitative Estimate of Drug-likeness. We primarily evaluate performance using \textit{Root Mean Square Error (RMSE)}, where a lower value indicates higher precision in capturing the chemical properties of the molecule.
    \item \textbf{IUPAC2Mol (IUPAC to Molecule):} This cross-modal generation task requires translating a formal chemical name into a molecular textual representation (SMILES or SELFIES). We utilize the \textit{Morgan Fingerprints Similarity (Morgan)} as the primary metric for generation quality. 
    \item \textbf{Mol2IUPAC (Molecule to IUPAC):} This task involves generating a standardized IUPAC name from a molecular textual representation. Performance is quantified using \textit{BLEU-4}, measuring the n-gram overlap between the generated nomenclature and the reference name.
\end{itemize}
Additionally, we calculate the performance degradation by:
\begin{equation}
\mathrm{Degradation}(k) = \sigma \cdot \frac{M(k) - M(\mathrm{train})}{M(\mathrm{train})} \times 100\%,
\end{equation}
where $\sigma = +1$ for lower-is-better metrics (RMSE) and $\sigma = -1$ for higher-is-better metrics (BLEU-4, Morgan similarity), and $M(\cdot)$ denotes the performance score.

\subsection{Task-wise Manifold Regularity}
Figures~\ref{fig:performance_ged} and \ref{fig:performance_heat} reveal a task-dependent robustness pattern under controlled structural perturbations, yet a common theme emerges across all three tasks: the region in which models behave reliably around in-distribution molecules is remarkably narrow, and even a few graph edits can push an input across a generalization boundary.

\noindent\textbf{Molecular understanding tasks exhibit steep early degradation with high model variance.}
For tasks that take molecular textual representations as input, e.g., \textbf{QED} prediction and \textbf{Mol2IUPAC}, we observe a characteristic two-phase pattern: a steep drop at low perturbation strength (GED=1--3) followed by a plateau or slower decline beyond GED=3. Even a single structural edit (GED=1) can already induce a non-trivial degradation across most models. While the maximal degradation at GED=5 remains moderate on average for Mol2IUPAC (up to $\sim$50\%), the steepness of the initial slope indicates that the model's predictive reliability is concentrated in an extremely narrow neighborhood around the original molecules. We further observe substantial cross-model variance on these understanding tasks: stronger domain-specific models might degrade more gracefully, whereas weaker general-purpose models such as Qwen3-1.7B exhibit catastrophic degradation (up to 175\% on QED at GED=5). This wide spread suggests that the size of the local ``trust region'' is highly sensitive to model capacity and pretraining choices, and that surface-level performance on the unperturbed test set can mask large differences in robustness. We further note that Mol2IUPAC is an inherently difficult task with limited training data, which likely amplifies the steepness of the initial degradation as models have not fully internalized the underlying mapping.

\noindent\textbf{Molecular generation exhibits a more consistent but pronounced collapse.}
In contrast, \textbf{IUPAC2Mol}, which requires generating a valid molecular structure from a textual description, shows a more uniform degradation pattern across models, with curves that decline consistently and flatten around GED=3, reaching a maximal degradation of approximately 60\% at GED=5. The relative consistency across models, together with the comparatively larger plateau magnitude, suggests that the bottleneck here is intrinsic to the task rather than to specific model choices: structure generation requires simultaneously satisfying syntactic validity (SMILES/SELFIES) and recovering precise connectivity, atom identities, bond orders, and branching. Consequently, the mapping between nearby input descriptions and output token sequences is highly non-smooth: small topological deviations in the implied target can trigger disproportionate changes in the correct output sequence.

\begin{wrapfigure}{r}{0.5\linewidth}
    \centering
    \vskip -0.3in
    \includegraphics[width=\linewidth]{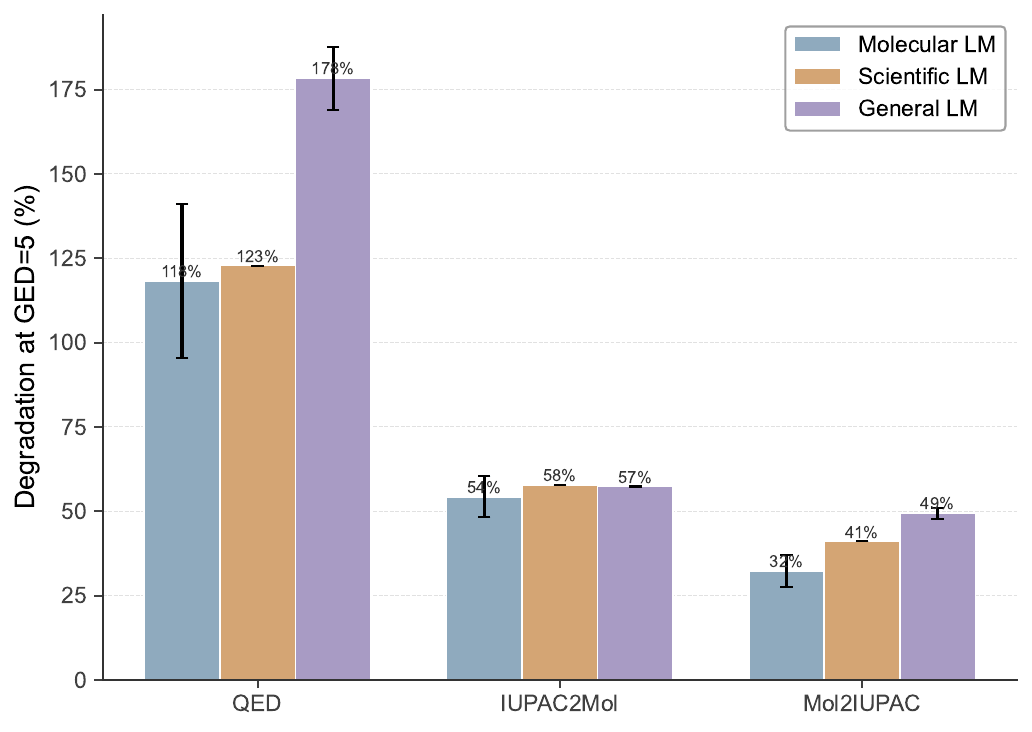}
    \vskip -0.1in
    \caption{Comparison of model pre-training strategy on extreme perturbation scenario (GED=5).}
    \vskip -0.2in
    \label{fig:arch}
\end{wrapfigure}

\subsection{Model-wise Manifold Regularity}

To examine how architectural and pretraining choices shape robustness, we further group the evaluated models into three categories: \emph{Molecular LLMs} (encoder--decoder models pretrained on chemical corpora, including MolT5 and BioT5), \emph{Scientific LLMs} (Galactica-125M), and \emph{General LLMs} (the Qwen3 series).

\noindent\textbf{Domain pretraining is critical for molecular understanding under perturbation.}
As shown in Figure~\ref{fig:arch}, the General Qwen3 models are markedly more susceptible to structural perturbations: under the extreme perturbation setting (GED=5), they consistently underperform the other two categories across all three tasks. The gap is most striking on QED prediction, where every Qwen3 variant suffers a degradation exceeding 150\%. This sharp contrast highlights the importance of domain-specific pretraining for molecular understanding tasks: without exposure to chemical corpora, general-purpose models fail to develop the local smoothness needed to remain stable under minor structural edits.

\noindent\textbf{Encoder-decoder Molecular LLMs are the most robust under extreme perturbations.}
Among the three categories, encoder-decoder Molecular LLMs exhibit the smallest performance degradation at GED=5 across all three tasks. The advantage is particularly pronounced on molecular understanding tasks: BioT5-base achieves the best extreme-perturbation stability on both QED and Mol2IUPAC, outperforming both Scientific and General LLMs by a wide margin. This suggests that the combination of encoder-decoder architecture and domain pretraining can learn a more smooth molecular representation space.

\noindent\textbf{Decoder-only models trade robustness for generative performance.
}
While encoder-decoder Molecular LLMs degrade more gracefully at GED=5, they do not always achieve the highest absolute performance. On the two generative tasks (IUPAC2Mol and Mol2IUPAC), decoder-only models attain stronger absolute scores on the unperturbed test set, despite suffering steeper relative drops under perturbation. We attribute this to the strong auto-regressive modeling capacity of decoder-only architectures in fitting the training distribution and generating long sequences.

\subsection{Effects of Perturbation Type} 
\begin{wrapfigure}{r}{0.5\linewidth}
    \centering
    \vskip -0.25in
    \includegraphics[width=\linewidth]{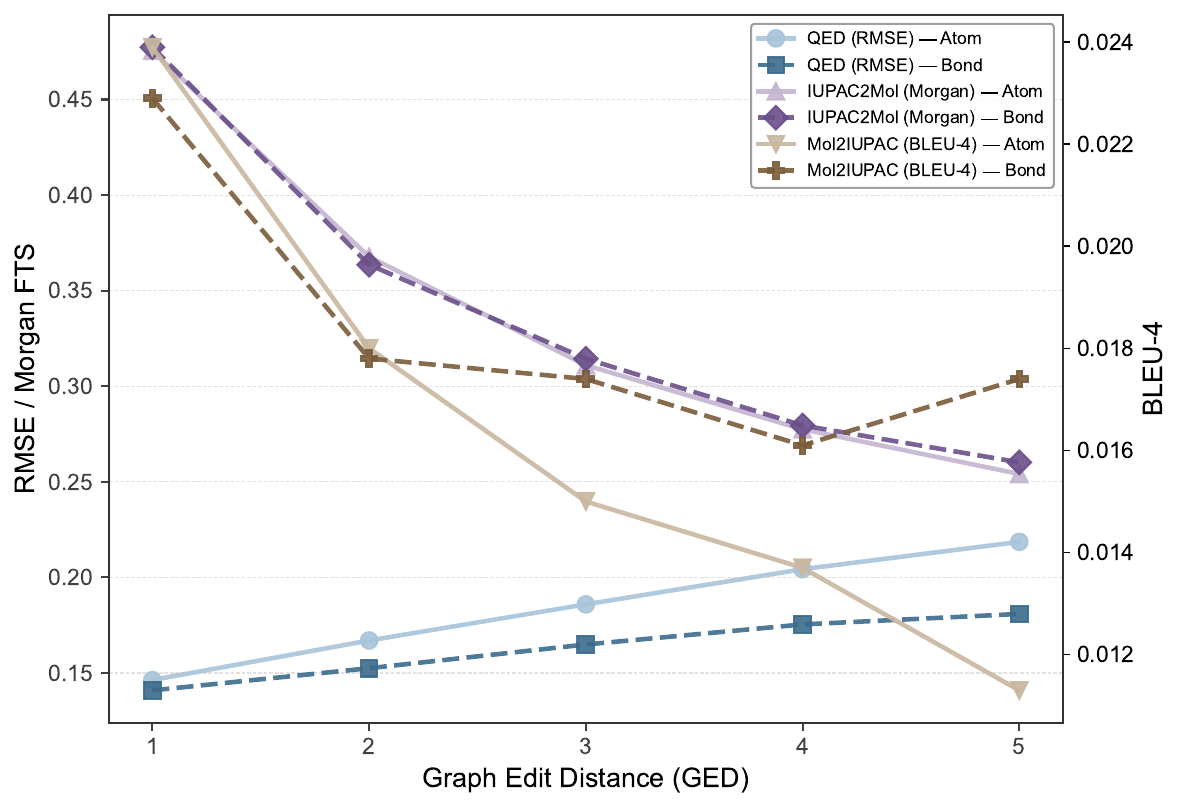}
    \vskip -0.1in
    \caption{Absolute performance trend of Galactica-125M corresponding to GED levels under atom-only and bond-only perturbations.}
    \vskip -0.15in
    \label{fig:atom_bond}
\end{wrapfigure}

We further disentangle the contributions of atom-level and bond-level perturbations by evaluating Galactica-125M under each type in isolation (Figure~\ref{fig:atom_bond}). The two perturbation types exhibit different effects depending on whether the task is generative or understanding-oriented.

For \textbf{IUPAC2Mol}, \emph{atom-only} and \emph{bond-only} perturbations induce nearly indistinguishable degradation curves across all GED levels. This symmetry is consistent with the nature of structure generation: the model must recover the exact connectivity and atom identities of the target molecule, so any modification, whether to a node or an edge, propagates through the output token sequence with comparable severity.

In contrast, on the two understanding tasks (\textbf{Mol2IUPAC} and \textbf{QED}), bond perturbations cause noticeably smaller degradation than atom perturbations at matched GED levels. We attribute this asymmetry to the task-specific invariances of each objective. QED is largely driven by the presence and arrangement of salient functional groups and global molecular descriptors (molecular weight, logP, hydrogen-bond donors/acceptors), most of which are determined primarily by atomic composition and only secondarily by exact bond patterns. Similarly, IUPAC names are organized around principal chains and substituent atoms: many bond modifications (e.g., saturation changes within a ring) leave the parent name skeleton unchanged or alter only a localized prefix, while atom substitutions typically force a more global renaming.

\section{In-Context Tuning under Molecular Perturbation}

\subsection{Theoretical Analysis}

\noindent\textbf{Setup and notation.}
Let $\mathcal{D}_{\text{train}} = \{(x_i, y_i)\}_{i=1}^{N} \subset \mathcal{X} \times \mathcal{Y}$ 
denote the training set, where the structure space $\mathcal{X}$ is equipped with a 
structural distance $d_\mathcal{X}: \mathcal{X} \times \mathcal{X} \to \mathbb{R}_{\ge 0}$ 
(e.g., graph edit distance), and the property space $\mathcal{Y}$ is a normed vector 
space with norm $\|\cdot\|$. For a query $\tilde{x}\in\mathcal{X}$, let 
$\mathcal{C}(\tilde{x}) \subseteq \mathcal{D}_{\text{train}}$ denote the retrieved 
context, consisting of its $n$ nearest neighbors under $d_\mathcal{X}$, written 
$\mathcal{C}(\tilde{x}) = \{(x_i, y_i)\}_{i=1}^{n}$. With slight abuse of notation, 
we write $d_\mathcal{X}(\tilde{x}, \mathcal{D}_{\text{train}}) := \min_{(x,y)\in\mathcal{D}_{\text{train}}} d_\mathcal{X}(\tilde{x}, x)$ 
for the distance from $\tilde{x}$ to its nearest training molecule.

\noindent\textbf{Assumption 1 (Lipschitz continuity).}
We assume the ground-truth property mapping $h^*: \mathcal{X} \to \mathcal{Y}$ is 
$L$-Lipschitz with respect to $d_\mathcal{X}$:
\begin{align}
    \| h^*(x_1) - h^*(x_2) \| \le L \cdot d_\mathcal{X}(x_1, x_2), 
    \quad \forall x_1, x_2 \in \mathcal{X}.
\end{align}
This is a direct formalization of the molecular similarity principle that ICT 
itself presupposes: structurally similar molecules carry similar properties, with 
$L$ quantifying the maximal sensitivity of $h^*$ to structural change. For tasks 
with discrete outputs, the assumption is read as a 
smoothness condition on a suitable divergence between output distributions; we 
develop the analysis in the normed-space setting for clarity and treat its 
qualitative implications as guidance for generative tasks.

\noindent\textbf{Off-manifold inference is not constrained by the training objective.}
Let $f_\theta: \mathcal{X} \to \mathcal{Y}$ be a model trained without context. 
For a query $\tilde{x}$ outside the training support 
(i.e., $d_\mathcal{X}(\tilde{x}, \mathcal{D}_{\text{train}}) > \delta$ for some 
$\delta>0$), the supervised objective imposes no constraint that ties 
$\|f_\theta(\tilde{x}) - h^*(\tilde{x})\|$ to $d_\mathcal{X}(\tilde{x}, \mathcal{D}_{\text{train}})$: 
whatever smoothness $f_\theta$ exhibits off-distribution must come from implicit 
regularization or architectural priors, neither of which is structurally aligned 
with the chemical metric $d_\mathcal{X}$. Consequently, errors are free to grow 
without a controlled dependence on the perturbation distance, consistent with 
the divergent degradation observed as GED increases.

\noindent\textbf{Local regularization via in-context inference.}
To gain analytical traction, we adopt an \emph{idealized} model of the in-context 
predictor. Let $g_\theta: \mathcal{X} \times 2^{\mathcal{X}\times\mathcal{Y}} \to \mathcal{Y}$ 
denote the in-context predictor that maps a query $\tilde{x}$ together with a 
context $\mathcal{C}(\tilde{x})$ to an output. Following a line of work that views 
in-context learning as performing implicit kernel regression~\citep{xie2021explanation,garg2022can}, 
we approximate $g_\theta$ by a softmax kernel smoother over the retrieved context labels:
\begin{equation}
    g_\theta\bigl(\tilde{x}, \mathcal{C}(\tilde{x})\bigr) 
    \approx \hat{y}_{\text{ict}} := \sum_{i=1}^n \alpha_i(\tilde{x};\theta)\, y_i, 
    \quad \alpha_i \ge 0, \; \sum_{i=1}^n \alpha_i = 1,
\end{equation}
where the weights $\alpha_i(\tilde{x};\theta)$ are induced by the trained attention 
mechanism. We treat this as a stylized abstraction rather than a literal 
description of attention dynamics; the gap between $g_\theta$ and its 
kernel-smoother form is absorbed into an additive approximation error 
$\varepsilon_{\text{approx}}$ that we do not attempt to bound. The purpose of the 
analysis below is not to predict ICT's error quantitatively, but to identify the 
geometric quantity that the retrieval procedure can directly control.

Let $R := \max_{(x_i, y_i) \in \mathcal{C}(\tilde{x})} d_\mathcal{X}(x_i, \tilde{x})$ 
denote the retrieval radius. Assuming clean labels $y_i = h^*(x_i)$, the triangle 
inequality combined with $L$-Lipschitzness yields:
\begin{align}
    \| \hat{y}_{\text{ict}} - h^*(\tilde{x}) \| 
    = \biggl\| \sum_{i=1}^n \alpha_i \bigl(h^*(x_i) - h^*(\tilde{x})\bigr) \biggr\|
    \le \sum_{i=1}^n \alpha_i \, \| h^*(x_i) - h^*(\tilde{x}) \|
    \le L \cdot R.
\end{align}
Combining with the approximation error gives the full bound 
$\|g_\theta(\tilde{x},\mathcal{C}(\tilde{x})) - h^*(\tilde{x})\| 
\le L\cdot R + \varepsilon_{\text{approx}}$.

\noindent\textbf{Proposition 1 (Locally bounded ICT error, idealized).}
\textit{Under the $L$-Lipschitz assumption on $h^*$ and the kernel-smoother 
idealization of $g_\theta$, the in-context prediction error satisfies}
\begin{align*}
    \bigl\|g_\theta\bigl(\tilde{x}, \mathcal{C}(\tilde{x})\bigr) - h^*(\tilde{x})\bigr\| 
    \le L\cdot R + \varepsilon_{\text{approx}},
\end{align*}
\textit{where $R$ is the retrieval radius. The bound is governed by the local 
geometric quantity $R$ rather than by the global perturbation distance 
$d_\mathcal{X}(\tilde{x}, \mathcal{D}_{\text{train}})$.}

Whereas the context-free error has no controlled dependence on 
$d_\mathcal{X}(\tilde{x}, \mathcal{D}_{\text{train}})$ and grows freely as 
$\tilde{x}$ moves off-manifold, the in-context error is constrained locally: by 
selecting nearest neighbors, the retrieval procedure directly minimizes the 
geometric quantity $R$ that governs the bound, confining the model to operate 
within a \emph{trust region} defined by the retrieved context. Two predictions 
follow that we will probe empirically: (i) replacing nearest-neighbor retrieval 
with random retrieval enlarges $R$ and should erode the bound; (ii) at large 
perturbation distances, even the nearest retrievable neighbor becomes far from 
$\tilde{x}$, so $R$ grows and the bound loosens.



\subsection{ICT and Retrieval-Based Control}

Standard Supervised Fine-Tuning (SFT) learns a direct mapping $f: x \to y$ by 
minimizing the negative log-likelihood over training pairs:
\begin{align}
    \mathcal{L}_{\text{SFT}}(\theta) 
    = -\mathbb{E}_{(x,y)\sim\mathcal{D}_{\text{train}}}\log p_\theta(y \mid x),
\end{align}
treating each sample in isolation without leveraging the local manifold structure 
of chemical space. In-Context Tuning (ICT) instead fine-tunes the model on $(x, y)$ 
pairs augmented with a retrieved context $C_x \subseteq \mathcal{D}_{\text{train}}\setminus\{x\}$ 
of the nearest structural neighbors of $x$. 
A vanilla in-context objective $-\log p_\theta(y \mid x, C_x)$ exposes the model 
to retrieved neighbors but does not require it to use them: the model may bypass 
$C_x$ and rely on its parametric prior, in which case the local-anchoring 
intuition of Proposition~1 fails to materialize at training time. 
We therefore add an auxiliary term to recover the in-context label mappings $x_i \to y_i$, aligning training with the kernel-smoothing primitive.
We expose the model to each training molecule under two complementary views via a Bernoulli indicator $v \in \{0, 1\}$: with probability $\tfrac{1}{2}$ ($v=0$) it sees $x$ alone; with probability $\tfrac{1}{2}$ ($v=1$) it sees $x$ together with $C_x$ and additionally recovers the in-context mappings. The combined objective is:
\begin{align}
    \mathcal{L}_{\text{ICT}}(\theta) = -\mathbb{E}_{(x,y),\, v\sim\text{Bern}(\tfrac{1}{2})}\!\bigl[\, \log p_\theta\bigl(y \mid x,\; v \cdot C_x\bigr) + v \cdot \!\!\sum_{(x_i,y_i)\in C_x}\!\!\log p_\theta(y_i \mid x_i) \,\bigr],
\end{align}
where $C_x$ retrieves the nearest structural neighbor of $x$ from the training set.

\begin{table}[t]
\centering
\vskip-0.3in
\resizebox{\textwidth}{!}{%
\begin{tabular}{llcccccc}
\toprule
\rowcolor{headergray}
\textbf{Task} & \textbf{Strategy} & \textbf{Raw Test} & $k{=}1$ & $k{=}2$ & $k{=}3$ & $k{=}4$ & $k{=}5$ \\
\midrule
\multirow{3}{*}{QED (RMSE$\downarrow$)} 
 & Direct     & 0.0915 & 0.1435 & 0.1602 & 0.1758 & 0.1914 & 0.2039 \\
 & ICT        & \textbf{0.0746} & \textbf{0.1272} & \textbf{0.1545} & 0.1756 & 0.1951 & 0.2075 \\
 & Random ICT & 0.1005 & 0.1447 & 0.1613 & \textbf{0.1752} & \textbf{0.1883} & \textbf{0.1945} \\
\midrule
\multirow{3}{*}{Mol2IUPAC (BLEU-4$\uparrow$)} 
 & Direct     & 0.7231 & 0.4887 & 0.4556 & 0.4427 & 0.4338 & 0.4263 \\
 & ICT        & \textbf{0.7635} & \textbf{0.5285} & \textbf{0.4959} & \textbf{0.4736} & \textbf{0.4604} & \textbf{0.4512} \\
 & Random ICT & 0.7096 & 0.4692 & 0.4534 & 0.4420 & 0.4297 & 0.4262 \\
\midrule
\multirow{3}{*}{IUPAC2Mol (Morgan$\uparrow$)} 
 & Direct     & 0.7092 & 0.5213 & 0.4148 & 0.3612 & 0.3237 & \textbf{0.3000} \\
 & ICT        & \textbf{0.7304} & \textbf{0.5573} & \textbf{0.4342} & \textbf{0.3676} & \textbf{0.3257} & 0.2977 \\
 & Random ICT & 0.3013 & 0.2325 & 0.2053 & 0.1916 & 0.1844 & 0.1798 \\
\bottomrule
\end{tabular}%
}
\caption{Empirical evaluation of in-context tuning under structural perturbation 
(Galactica-125M). Best results in \textbf{bold}.}
\label{tab:main}
\vskip-0.3in
\end{table}

\subsection{Findings}

We evaluate three conditions on Galactica-125M across GED-perturbed inputs 
($k\in\{1,\dots,5\}$) with context size $n=1$: standard fine-tuning 
(\textbf{Direct}), \textbf{ICT} with nearest-neighbor retrieval, and 
\textbf{Random ICT}, which retains the in-context objective but draws context 
uniformly at random from the training set, isolating the role of structural 
similarity from in-context conditioning itself.

\noindent\textbf{ICT confers consistent but task-dependent robustness against structural perturbation.}
ICT improves over Direct in $15$ of $18$ task--perturbation cells in 
Table~\ref{tab:main}. On Mol2IUPAC, ICT dominates Direct at every level, with 
BLEU-4 gains of $\approx 0.04$ from raw to $k=2$ and $0.025$--$0.03$ at higher 
perturbation. On IUPAC2Mol, ICT wins at five of six levels, narrowing from a 
$0.036$ lead at $k=1$ to a near-tie at $k=5$ ($0.2977$ vs.\ $0.3000$). On QED, 
ICT improves clean RMSE substantially ($0.0746$ vs.\ $0.0915$) and remains 
better through $k=2$, but its advantage erodes at $k\geq 3$. ICT thus extends 
the local trust region in the majority of conditions, but the magnitude of the 
benefit varies markedly with task and perturbation level.

\noindent\textbf{Retrieval quality is essential: random neighbors largely eliminate the benefit.}
Random ICT isolates structural similarity from the mere presence of context. 
On IUPAC2Mol, random retrieval is catastrophic: Morgan FTS collapses from 
$0.7092$ (Direct) to $0.3013$ at clean and stays below $0.25$ throughout. 
On Mol2IUPAC, Random ICT is consistently worse than ICT and slightly worse 
than Direct. These patterns are consistent with Proposition~1: random retrieval 
enlarges $R$ and dissolves the locally bounded error guarantee. The exception 
is QED, where Random ICT is competitive and best at $k\ge 3$; we attribute this 
to a regression-specific mean-regression effect on a bounded $[0,1]$ target. 
Overall, the benefit of ICT derives specifically from \emph{structural} 
similarity.

\noindent\textbf{Limitations of ICT-based Anchoring.}
The robustness gains of ICT diminish as perturbation grows: on Mol2IUPAC the 
gap to Direct narrows from $0.040$ at $k=1$ to $0.025$ at $k=5$; on IUPAC2Mol 
from $0.036$ at $k=1$ to a tie at $k=5$; on QED, ICT loses its lead entirely 
by $k=4$. This shared trend points to retrieval quality, rather than the 
training objective, as the dominant factor at large $k$. When the perturbed 
query has drifted far from the training manifold, the nearest retrievable 
neighbor is no longer truly nearby: $R$ in Proposition~1 grows, the local-anchor 
argument loosens, and the context can even induce negative transfer if it 
belongs to a different region of chemical space. ICT's effectiveness is 
therefore fundamentally bounded by the geometry of $\mathcal{D}_{\text{train}}$: 
it stabilizes predictions inside a moderate neighborhood of the training 
distribution, but cannot manufacture local structure where none exists.
\section{Conclusion}
We introduce a syntax-valid Molecular Perturbation framework to probe the 
manifold regularity of molecular LLMs via controlled structural edits (GED). 
Our results indicate a narrow local \emph{trust region}: even a single graph 
edit can induce substantial performance degradation, highlighting limited 
generalization beyond the training distribution. Motivated by the molecular 
similarity principle, we study whether In-Context Tuning (ICT), which anchors 
predictions on structurally similar retrieved molecules, can mitigate this 
fragility. ICT consistently improves robustness within a local neighborhood, 
but its gains diminish as queries move further off-manifold. Overall, these 
findings suggest that retrieval-based anchoring is a promising yet inherently 
local mechanism, and underscore the need for evaluation and training strategies 
that account for the topological sensitivity of chemical space.

\bibliography{reference}
\bibliographystyle{rusnat}

\newpage
\appendix
\definecolor{example_back}{rgb}{1,1,1}
\definecolor{example_front}{rgb}{0.8, 0.8, 0.8}
\definecolor{prompt_front}{rgb}{0.859, 0.918, 0.922}
\definecolor{prompt_back}{rgb}{0.953, 0.971, 0.976}
\definecolor{human_front}{rgb}{0.784, 0.686, 0.604}
\definecolor{human_back}{rgb}{1,1,1}
\definecolor{case_front}{rgb}{1,1,1}
\definecolor{case_back}{rgb}{1,1,1}

\section{Detailed Experiment Setup \& Hyper Parameters}
\label{sec:appendix_setup}

\subsection{Hyper-parameters}
Table~\ref{tab:hyperparams} summarizes the default hyper-parameters used in training and inference.
We report the main settings (e.g., batch size, learning rate, and decoding configuration) for reproducibility.
All experiments were conducted on NVIDIA H200 GPUs with 141\,GB memory.
Additionally, all models are \emph{fully fine-tuned} (no parameter-efficient adapters such as LoRA) on the training data.

\begin{table}[htbp]
    \centering
    \resizebox{0.4\textwidth}{!}{%
    \begin{tabular}{ll}
        \toprule
        \rowcolor{headergray}
        \textbf{Item} & \textbf{Value} \\
        \midrule
        \emph{Training (Decoder-only)}\\
        Batch size & 32 \\
        Micro batch size & 4 \\
        Gradient accumulation steps & 8 \\
        Epochs & 10 \\
        Warm-up steps & 200 \\
        Learning rate & $2 \times 10^{-4}$ \\
        Optimizer & AdamW \\
        Cutoff length & 512 \\
        Precision & FP32 \\
        \hline
        \emph{Training (Encoder-decoder)}\\
        Micro batch size & 8 \\
        Epochs & 10 \\
        Warm-up steps & 200 \\
        Learning rate & $2 \times 10^{-4}$ \\
        LR scheduler & Cosine \\
        Cutoff length & 512 \\
        \hline
        \emph{Inference (Decoder-only)}\\
        Temperature & 0.7 \\
        Top-$p$ & 0.85 \\
        Top-$k$ & 40 \\
        Max new tokens & 256 \\
        \hline
        \emph{Inference (Encoder-decoder)}\\
        Decoding & Greedy \\
        Max new tokens & 512 \\
        \bottomrule
    \end{tabular}
    }
    \caption{Hyper-parameters.}
    \label{tab:hyperparams}
\end{table}

\subsection{Perturbed Data Statistics}

\begin{table}[t]
\centering
\resizebox{0.8\columnwidth}{!}{
\begin{tabular}{cccc}
\toprule
\rowcolor{headergray}
\textbf{Dataset} & \textbf{Total Samples} & \textbf{Atom-only Perturbations} & \textbf{Bond-only Perturbations} \\
\midrule
raw\_train & 11853 &N/A & N/A \\
raw\_test & 1985 &N/A & N/A \\
k=1 & 16,162 & 7,838 & 8,324 \\
k=2 & 18,899 & 9,366 & 9,533 \\
k=3 & 18,485 & 9,326 & 9,159 \\
k=4 & 17,368 & 9,317 & 8,051 \\
k=5 & 15,633 & 9,184 & 6,449 \\
\bottomrule
\end{tabular}
}
\caption{Statistics of the raw and perturbed datasets.}
\vskip -0.2in
\label{tab:perturb-stats}
\end{table}

We construct our evaluation dataset by applying controlled perturbations to PubChem \citep{liu2023molca}. For each molecule in the original training set (i.e., \textit{raw\_train}), we generate variants across the five discrete levels of Graph Edit Distance (GED). Crucially, we isolate perturbations by applying either atom-only or bond-only modifications. This controlled separation allows us to disentangle \emph{node} (atom identity) versus \emph{edge} (bond order) sensitivity, and thereby directly test the manifold-regularity hypothesis that whether small, syntax valid topological edits lead to smooth and predictable changes in model behavior, or instead push the model across a generalization boundary.

Table~\ref{tab:perturb-stats} summarizes the statistics of the original and perturbed datasets. 
For each GED level $k$, we report the total number of valid perturbed molecules, along with a breakdown of variants induced by atom-only or bond-only perturbations.
Notably, as GED increases, the proportion of bond-level perturbations gradually decreases, reflecting the stricter syntax validity constraints associated with bond order modifications. 
In contrast, atom-level substitutions remain relatively stable across GED levels, indicating that atomic edits are more frequently admissible under the enforced valence constraints.

\begin{table}[t]
\centering
\small
\begin{tabular}{lccc}
\toprule
\rowcolor{headergray}\textbf{Atom} & \textbf{Raw Frequency} & \textbf{Perturbed Frequency} & $\Delta$ \\
\midrule
C  & 68.1\% & 64.3\% & $-3.8\%$ \\
O  & 22.8\% & 22.8\% & $0.0\%$ \\
N  & 6.2\%  & 6.9\%  & $+0.7\%$ \\
S  & 0.8\%  & 1.5\%  & $+0.7\%$ \\
P  & 0.8\%  & 1.4\%  & $+0.6\%$ \\
Cl & 0.5\%  & 1.0\%  & $+0.5\%$ \\
F  & 0.3\%  & 0.9\%  & $+0.6\%$ \\
B  & $<0.1\%$ & 0.8\% & $+0.8\%$ \\
\bottomrule
\end{tabular}
\caption{Atom frequency distribution before and after perturbation. $\Delta$ denotes the change from raw to perturbed.}
\label{tab:atom_frequency}
\end{table}

A natural concern with atom-level perturbations is whether they push molecules into chemically implausible regions. To verify that our perturbations remain chemically meaningful, we compare the atom-frequency distributions of the raw and perturbed corpora, shown in Table~\ref{tab:atom_frequency}. The two distributions are highly consistent: carbon and oxygen, which together account for over 90\% of all atoms in the raw data, remain the dominant elements after perturbation, with only a marginal $-3.8\%$ shift in carbon frequency. Other common organic atoms (N, S, P, Cl, F) see small absolute increases of $0.5$--$0.7\%$, all within the range of typical drug-like molecular composition. Although rare atoms such as boron exhibit relatively larger fold-changes due to their near-zero baseline frequency, their absolute share remains below $1\%$. Overall, the perturbed corpus retains the elemental signature of the raw training distribution, confirming that our perturbations probe local neighborhoods within the realistic chemical space.

\subsection{Similarity Distributions Under Perturbations}
\begin{figure*}[ht]
    \centering
    \includegraphics[width=1.0\linewidth]{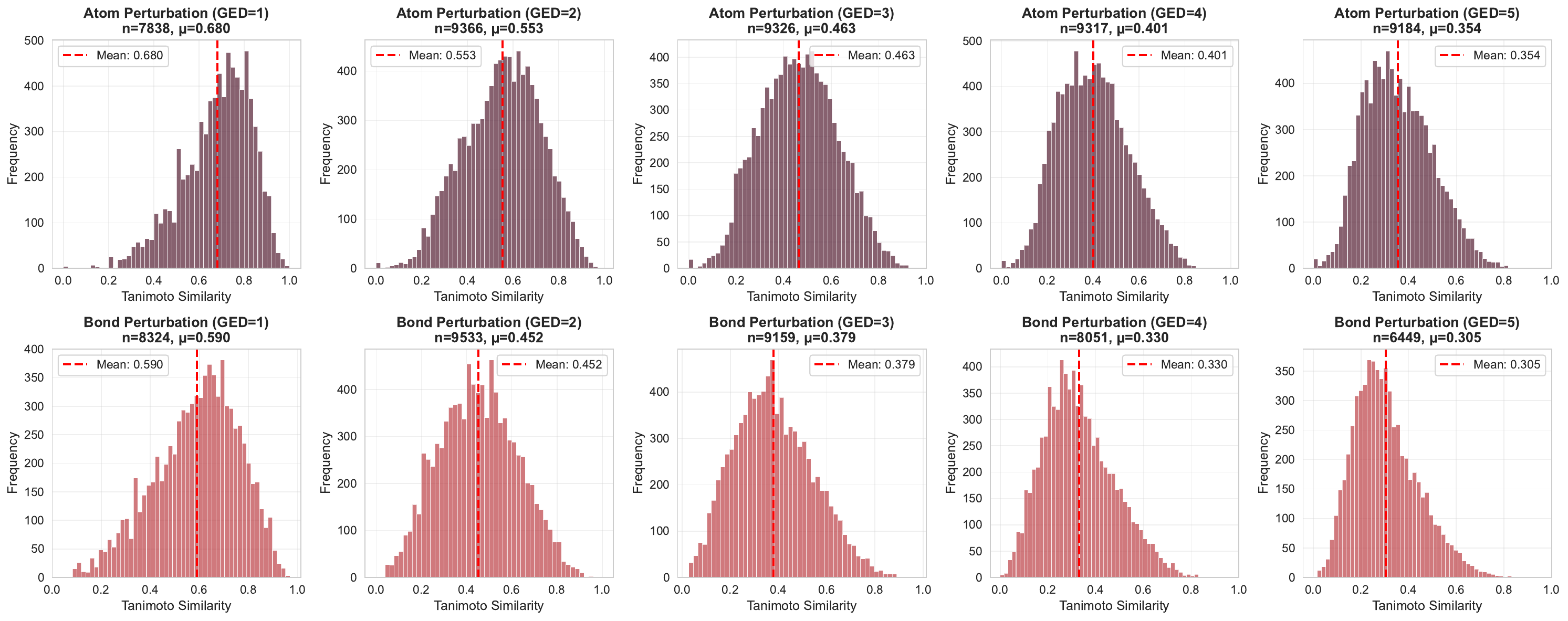}
    \vskip -0.1in
    \caption{Histograms of Tanimoto similarities between original and perturbed molecules across GED levels from 1 to 5 (Morgan fingerprints, radius 2, 2048 bits). The top row corresponds to atom-level perturbations and the bottom row to bond-level perturbations; each panel reports the sample size $n$ and mean similarity $\mu$.}
    \label{fig:dist}
\end{figure*}

To quantify how much a perturbed molecule deviates from its original counterpart, we compute Tanimoto similarity between Morgan fingerprints (radius 2, 2048 bits) for both atom- and bond-level perturbations across GED levels. Figure~\ref{fig:dist} visualizes the resulting similarity histograms (top: atom perturbations; bottom: bond perturbations).

Across both perturbation types, similarity decreases monotonically as GED ($k$) increases, indicating progressively larger structural changes. For atom perturbations, the mean similarity drops from 0.68 (k=1, $n=7{,}838$) to 0.55 (k=2, $n=9{,}366$), 0.46 (k=3, $n=9{,}326$), 0.40 (k=4, $n=9{,}317$), and 0.35 (k=5, $n=9{,}184$). Bond perturbations exhibit the same trend but yield consistently lower similarities at matched k, with means of 0.59 (k=1, $n=8{,}324$), 0.45 (k=2, $n=9{,}533$), 0.38 (k=3, $n=9{,}159$), 0.33 (k=4, $n=8{,}051$), and 0.30 (k=5, $n=6{,}449$). Overall, these distributions confirm that (i) larger GED corresponds to lower molecular similarity and (ii) bond edits tend to induce larger deviations than atom edits under the same GED level.

\subsection{Retrieval Setting}

For tasks whose input is a SMILES string (Mol2IUPAC, QED, LogP, HOMO-LUMO gap, polarizability), we retrieve the nearest training neighbor using \textbf{Morgan fingerprints} (radius\,=\,2, 2048 bits) with Tanimoto similarity.
For IUPAC2Mol, where the input is an IUPAC name, we use \textbf{BM25} over tokenized IUPAC names to obtain the nearest neighbor.


\section{Performance of Additional LLMs}

\begin{table}[t]
\centering
\small
\resizebox{0.85\columnwidth}{!}{
\begin{tabular}{llcccccc}
\toprule
\rowcolor{headergray}
\textbf{Model} & \textbf{Task (Metric)} & \textbf{Raw} & $k{=}1$ & $k{=}2$ & $k{=}3$ & $k{=}4$ & $k{=}5$ \\
\midrule
\multirow{3}{*}{LLaMA-3.2-1B}
 & QED (RMSE $\downarrow$)         & 0.1302 & 0.1600 & 0.1733 & 0.1857 & 0.1999 & 0.2119 \\
 & Mol2IUPAC (BLEU-4 $\uparrow$)     & 0.3843 & 0.2500 & 0.2244 & 0.2192 & 0.2120 & 0.2098 \\
 & IUPAC2Mol (Morgan FTS $\uparrow$) & 0.4611 & 0.3629 & 0.3069 & 0.2764 & 0.2515 & 0.2368 \\
\midrule
\multirow{3}{*}{ChemLLM-7B}
 & QED (RMSE $\downarrow$)         & 0.2677 & 0.2456 & 0.1957 & 0.2936 & 0.2788 & 0.2457 \\
 & Mol2IUPAC (BLEU-4 $\uparrow$)     & 0.4268 & 0.1991 & 0.1797 & 0.1743 & 0.1690 & 0.1658 \\
 & IUPAC2Mol (Morgan FTS $\uparrow$) & 0.5108 & 0.3869 & 0.3247 & 0.2930 & 0.2685 & 0.2524 \\
\bottomrule
\end{tabular}
}
\caption{Performance of LLaMA-3.2-1B and ChemLLM-7B under structural perturbations of increasing GED ($k=1,\dots,5$) using standard fine-tuning. $\uparrow$/$\downarrow$ indicates higher/lower is better.}
\label{tab:additional_models}
\end{table}
To verify that our findings generalize beyond the models in the main analysis, we further evaluate two additional architectures under the same perturbation protocol: LLaMA-3.2-1B (a general-purpose LLM) and ChemLLM-7B (a chemistry-pretrained LLM). Table~\ref{tab:additional_models} reports their performance across the three tasks at GED levels $k=1,\dots,5$.

Both models reproduce the characteristic two-phase pattern observed earlier: a steep drop at low perturbation strength ($k=1$) followed by a slower decline as $k$ grows. On the two generative tasks, the trend is particularly pronounced: Mol2IUPAC BLEU-4 falls by roughly $35\%$ for LLaMA-3.2-1B and over $50\%$ for ChemLLM-7B at $k=1$ alone compared to the raw test set, and IUPAC2Mol Morgan similarity exhibits a similarly steep initial slope before flattening. This consistency across model scales (1B vs.\ 7B) and pretraining regimes (general vs.\ chemistry-specific) reinforces our central finding: the narrow local trust region is a property of the molecular sequence-modeling paradigm itself, not an artifact of a particular model choice.

Interestingly, ChemLLM-7B, despite being seven times larger and pretrained on chemistry corpora, does not exhibit better robustness than the smaller LLaMA-3.2-1B. On Mol2IUPAC, ChemLLM-7B starts from a higher raw BLEU-4 ($0.4268$ vs.\ $0.3843$) but ends up below LLaMA-3.2-1B at every perturbation level ($0.1658$ vs.\ $0.2098$ at $k=5$), indicating a steeper relative drop. The QED column further shows a non-monotonic trajectory for ChemLLM-7B, where RMSE fluctuates rather than degrading smoothly with $k$. 

\section{Absolute Performance}
\label{app:full_results}
Tables~\ref{tab:qed_results},~\ref{tab:mol2cap_results}, and~\ref{tab:IUPAC2Mol_results} report the complete numerical results for all models across three downstream tasks under increasing structural perturbation levels (GED $k{=}1,\dots,5$).
\begin{table}[t]
\centering
\small
\resizebox{\columnwidth}{!}{
\begin{tabular}{llcccccc}
\toprule
\rowcolor{headergray}
\textbf{Architecture} & \textbf{Model} & \textbf{Raw Test} & $k{=}1$ & $k{=}2$ & $k{=}3$ & $k{=}4$ & $k{=}5$ \\
\midrule
\multirow{4}{*}{Encoder-Decoder}
 & MolT5-small   & 0.0804 & 0.1222 & 0.1423 & 0.1587 & 0.1736 & 0.1829 \\
 & MolT5-base   & 0.0646 & 0.1045 & 0.1146 & 0.1213 & 0.1279 & 0.1292 \\
 & MolT5-large  & 0.0547 & 0.0973 & 0.1116 & 0.1221 & 0.1313 & 0.1375 \\
 & BioT5-base   & 0.0713 & 0.1064 & 0.1207 & 0.1297 & 0.1366 & 0.1388 \\
\midrule
\multirow{6}{*}{Decoder-Only}
 & Galactica-125M      & 0.0915 & 0.1435 & 0.1602 & 0.1758 & 0.1914 & 0.2039 \\
 & Qwen3-0.6B          & 0.0656 & 0.1166 & 0.1326 & 0.1478 & 0.1620 & 0.1750 \\
 & Qwen3-1.7B          & 0.0660 & 0.1262 & 0.1458 & 0.1625 & 0.1791 & 0.1914 \\
 & Qwen3-4B            & 0.0925 & 0.1654 & 0.1830 & 0.2032 & 0.2208 & 0.2574 \\
 & LLaMA-3.2-1B        & 0.1302 & 0.1600 & 0.1733 & 0.1857 & 0.1999 & 0.2119 \\
 & ChemLLM-7B          & 0.2677 & 0.2456 & 0.1957 & 0.2936 & 0.2788 & 0.2457 \\
\bottomrule
\end{tabular}
}
\caption{Full results of property prediction (QED) performance measured by RMSE ($\downarrow$) under increasing GED levels.}
\label{tab:qed_results}
\end{table}

\begin{table}[t]
\centering
\small
\resizebox{\columnwidth}{!}{
\begin{tabular}{llcccccc}
\toprule
\rowcolor{headergray}
\textbf{Architecture} & \textbf{Model} & \textbf{Raw Test} & $k{=}1$ & $k{=}2$ & $k{=}3$ & $k{=}4$ & $k{=}5$ \\
\midrule
\multirow{4}{*}{Encoder-Decoder}
 & MolT5-small   & 0.3464 & 0.2978 & 0.2798 & 0.2744 & 0.2660 & 0.2581 \\
 & MolT5-base   & 0.3656 & 0.2931 & 0.2710 & 0.2657 & 0.2590 & 0.2529 \\
 & MolT5-large  & 0.4558 & 0.3522 & 0.3267 & 0.3209 & 0.3104 & 0.3007 \\
 & BioT5-base   & 0.5969 & 0.4578 & 0.4152 & 0.3976 & 0.3788 & 0.3660 \\
\midrule
\multirow{6}{*}{Decoder-Only}
 & Galactica-125M      & 0.7231 & 0.4887 & 0.4556 & 0.4427 & 0.4338 & 0.4263 \\
 & Qwen3-0.6B          & 0.7551 & 0.4668 & 0.4357 & 0.4225 & 0.4084 & 0.3997 \\
 & Qwen3-1.7B          & 0.7649 & 0.4398 & 0.4096 & 0.3972 & 0.3841 & 0.3796 \\
 & Qwen3-4B            & 0.7711 & 0.4372 & 0.4130 & 0.3994 & 0.3867 & 0.3799 \\
 & LLaMA-3.2-1B        & 0.3843 & 0.2500 & 0.2244 & 0.2192 & 0.2120 & 0.2098 \\
 & ChemLLM-7B          & 0.4268 & 0.1991 & 0.1797 & 0.1743 & 0.1690 & 0.1658 \\
\bottomrule
\end{tabular}
}
\caption{Full results of Molecule to IUPAC (Mol2IUPAC) performance measured by BLEU-4 ($\uparrow$) under increasing GED levels.}
\label{tab:mol2cap_results}
\end{table}

\begin{table}[t]
\centering
\small
\resizebox{\columnwidth}{!}{
\begin{tabular}{llcccccc}
\toprule
\rowcolor{headergray}
\textbf{Architecture} & \textbf{Model} & \textbf{Raw Test} & $k{=}1$ & $k{=}2$ & $k{=}3$ & $k{=}4$ & $k{=}5$ \\
\midrule
\multirow{4}{*}{Encoder-Decoder}
 & MolT5-small   & 0.2261 & 0.1800 & 0.1554 & 0.1414 & 0.1286 & 0.1224 \\
 & MolT5-base   & 0.3496 & 0.2661 & 0.2246 & 0.2013 & 0.1811 & 0.1669 \\
 & MolT5-large  & 0.4854 & 0.3491 & 0.2846 & 0.2529 & 0.2263 & 0.2119 \\
 & BioT5-base   & 0.4760 & 0.2903 & 0.2216 & 0.1958 & 0.1916 & 0.1782 \\
\midrule
\multirow{6}{*}{Decoder-Only}
 & Galactica-125M      & 0.7092 & 0.5213 & 0.4148 & 0.3612 & 0.3237 & 0.3000 \\
 & Qwen3-0.6B          & 0.7372 & 0.5604 & 0.4419 & 0.3821 & 0.3405 & 0.3139 \\
 & Qwen3-1.7B          & 0.7535 & 0.5855 & 0.4597 & 0.3945 & 0.3499 & 0.3236 \\
 & Qwen3-4B            & 0.7604 & 0.5958 & 0.4681 & 0.3996 & 0.3534 & 0.3231 \\
 & LLaMA-3.2-1B        & 0.4611 & 0.3629 & 0.3069 & 0.2764 & 0.2515 & 0.2368 \\
 & ChemLLM-7B          & 0.5108 & 0.3869 & 0.3247 & 0.2930 & 0.2685 & 0.2524 \\
\bottomrule
\end{tabular}
}
\caption{Full results of IUPAC to Molecule (IUPAC2Mol) performance measured by Morgan FTS ($\uparrow$) under increasing GED levels.}
\label{tab:IUPAC2Mol_results}
\end{table}

\section{Performance on Additional Molecular Property Prediction Tasks}

\begin{table}[t]
\centering
\small
\resizebox{0.9\columnwidth}{!}{
\begin{tabular}{llccccccc}
\toprule
\rowcolor{headergray}
\textbf{Property} & \textbf{Strategy} & \textbf{Raw Test} & $k{=}1$ & $k{=}2$ & $k{=}3$ & $k{=}4$ & $k{=}5$ \\
\midrule
\multirow{2}{*}{LogP (MAE $\downarrow$)}
 & Direct & 0.5374 & 1.2576 & 1.3891 & 1.4540 & 1.5287 & 1.6063 \\
 & ICT    & \textbf{0.4824} & \textbf{0.8175} & \textbf{0.9688} & \textbf{1.1025} & \textbf{1.2475} & \textbf{1.4286} \\
\midrule
\multirow{2}{*}{HOMO-LUMO Gap (MAE $\downarrow$, eV)}
 & Direct & \textbf{0.3375} & 1.7622 & 1.9701 & 1.9486 & 2.0504 & 2.1374 \\
 & ICT    & 0.4434 & \textbf{1.4387} & \textbf{1.7028} & \textbf{1.6872} & \textbf{1.7352} & \textbf{1.8067} \\
\midrule
\multirow{2}{*}{Polarizability (MAE $\downarrow$)}
 & Direct & \textbf{4.3439} & 10.1769 & 10.9901 & 11.0478 & 11.6991 & 12.0652 \\
 & ICT    & 5.3884 & \textbf{7.2150} & \textbf{7.9970} & \textbf{8.2379} & \textbf{9.1079} & \textbf{9.4900} \\
\bottomrule
\end{tabular}
}
\caption{Performance of Galactica-125M on three additional molecular property prediction tasks (LogP, HOMO-LUMO Gap, Polarizability) under structural perturbations of increasing GED ($k=1,\dots,5$). \textbf{Bold} marks the best value per column. ICT uses a single retrieved neighbor ($n=1$).}
\label{tab:additional_properties}
\end{table}
To verify that our findings are not specific to QED, we evaluate ICT on three additional property prediction tasks: LogP (lipophilicity), HOMO-LUMO Gap (electronic structure), and Polarizability. Table~\ref{tab:additional_properties} reports MAE under standard fine-tuning (Direct) and ICT with a single retrieved neighbor.

The results closely mirror the pattern observed on QED in our main analysis. First, all three properties exhibit substantial degradation under perturbation: even a single graph edit ($k=1$) more than doubles the MAE relative to the unperturbed test set across all settings, confirming that the narrow local trust region we identify is a general property of regression tasks rather than a quirk of QED. Second, ICT consistently improves robustness across all three properties: it dominates Direct at every perturbation level ($k=1,\dots,5$), with particularly large gains on LogP ($1.4286$ vs.\ $1.6063$ at $k=5$, a relative reduction of $11\%$) and Polarizability ($9.4900$ vs.\ $12.0652$ at $k=5$, a $21\%$ reduction). Third, the same trade-off between clean and perturbed performance recurs: on HOMO-LUMO Gap and Polarizability, Direct retains a slight edge on raw test inputs, but ICT takes over as soon as any structural perturbation is applied. This consistent pattern across four independent properties demonstrates that ICT's local-anchoring mechanism generalizes broadly within molecular property prediction, and that the manifold-regularity issues we identify in the main paper are not idiosyncratic to a single property but reflect a structural property of the sequence-based modeling paradigm.

\section{ICT Performance on Encoder-decoder Models}

\begin{table}[t]
\centering
\small
\resizebox{0.85\columnwidth}{!}{
\begin{tabular}{llcccccc}
\toprule
\rowcolor{headergray}
\textbf{Task} & \textbf{Strategy} & \textbf{Raw Test} & $k{=}1$ & $k{=}2$ & $k{=}3$ & $k{=}4$ & $k{=}5$ \\
\midrule
\multirow{2}{*}{\shortstack[l]{Mol2IUPAC\\(BLEU-4 $\uparrow$)}}
 & Direct   & \textbf{0.7842} & 0.6031 & 0.5437 & \textbf{0.5217} & \textbf{0.5020} & \textbf{0.4923} \\
 & ICT & 0.7815 & \textbf{0.6045} & \textbf{0.5461} & 0.5208 & 0.5009 & 0.4894 \\
\midrule
\multirow{2}{*}{\shortstack[l]{IUPAC2Mol\\(Morgan $\uparrow$)}}
 & Direct   & 0.4492 & 0.2645 & 0.1810 & \textbf{0.1332} & \textbf{0.1607} & 0.1404 \\
 & ICT & \textbf{0.4506} & \textbf{0.2838} & \textbf{0.1827} & 0.1133 & 0.1490 & \textbf{0.1778} \\
\midrule
\multirow{2}{*}{\shortstack[l]{QED\\(RMSE $\downarrow$)}}
 & Direct   & \textbf{0.0713} & \textbf{0.1064} & \textbf{0.1207} & \textbf{0.1297} & \textbf{0.1366} & \textbf{0.1388} \\
 & ICT & 0.0759 & 0.1164 & 0.1300 & 0.1376 & 0.1453 & 0.1490 \\
\bottomrule
\end{tabular}
}
\caption{Performance comparison between Direct generation and ICT under increasing GED levels ($k=1,\dots,5$) on BioT5-base. Best results are \textbf{bold}.}
\label{tab:biot5_direct_vs_fullict}
\end{table}
Table~\ref{tab:biot5_direct_vs_fullict} compares Direct generation and ICT on BioT5-base across three representative tasks.
Unlike the consistent gains observed with Galactica-125M, ICT provides \emph{no clear advantage} on the encoder-decoder architecture.

For \textbf{Mol2IUPAC}, the two strategies are nearly indistinguishable: ICT edges ahead at $k{=}1$ and $k{=}2$ by less than 0.003 BLEU-4, while Direct takes a marginal lead from $k{=}3$ onward.
The degradation trajectories are also almost identical ($-$0.292 vs.\ $-$0.292 from raw to $k{=}5$), suggesting that the in-context example neither helps nor hurts when the encoder already captures sufficient SMILES-to-IUPAC correspondence.

For \textbf{IUPAC2Mol}, the picture is mixed.
ICT achieves higher Morgan fingerprint similarity at $k{=}1$ (+0.019) and $k{=}5$ (+0.037), but Direct is better at $k{=}3$ and $k{=}4$.
Both strategies suffer from extremely low SMILES validity ($<$2\%), indicating that BioT5-base struggles with syntactically correct SMILES generation regardless of the training strategy.
The most decisive gap appears in \textbf{QED prediction}, where Direct outperforms ICT at every perturbation level.
The RMSE advantage widens from 0.005 at raw to 0.010 at $k{=}5$, and the $R^2$ gap grows from 0.012 to 0.087.

More importantly, Direct degrades more gracefully: its $R^2$ drops by 0.481 (raw$\to$$k{=}5$) compared to 0.556 for ICT, a 15.6\% larger collapse.
This indicates that forcing all samples into ICT format---including those with low-similarity neighbors---injects noise that is particularly harmful for numerical regression.
These results highlight a key architectural difference: decoder-only models like Galactica benefit from in-context demonstrations even when neighbor quality is uncontrolled, whereas encoder-decoder models like BioT5 are more sensitive to noisy examples.
This observation motivates the similarity-threshold gating mechanism in our ICT strategy, which selectively applies ICT only when a sufficiently similar neighbor is available, thereby avoiding the noise penalty exposed by ICT.

\section{Related Work for LLM Generalization}
The generalization and robustness of large language models under input perturbation has been extensively studied in the NLP domain. \cite{kumar2025robustness} provide a comprehensive survey on LLM robustness, categorizing sources of non-robustness into intrinsic model limitations, data-driven vulnerabilities, and adversarial factors. Notably, even minimal input perturbations can cause significant performance degradation: \cite{gan2024reasoning} demonstrate that just 1 character edit on Mistral-7B-Instruct drops GSM8K accuracy from 43.7\% to 38.6\%, declining further to 19\% with 8 edits, a pattern strikingly analogous to our findings in the molecular domain. \cite{lou2024cr} further develop certified robustness methods against universal text perturbations using random smoothing, while \cite{yang2024assessing} systematically evaluate adversarial robustness across open-source LLMs including LLaMA, OPT, and T5, revealing that robustness varies substantially with model size and architecture. A recent comprehensive survey by \cite{zhang2026generalizability} formalizes the generalizability of LLM-based agents as the ability to maintain consistent performance across varied tasks and domains beyond fine-tuning data, proposing a hierarchical domain-task ontology for evaluation.

\definecolor{IUPAC2Mol_front}{RGB}{124, 138, 150}
\definecolor{IUPAC2Mol_back}{RGB}{248, 250, 251}
\definecolor{Mol2IUPAC_front}{RGB}{120, 140, 185}
\definecolor{Mol2IUPAC_back}{RGB}{246, 248, 253}
\definecolor{qed_front}{RGB}{125, 160, 130}
\definecolor{qed_back}{RGB}{246, 251, 247}

\begin{tcolorbox}[colframe=IUPAC2Mol_front, colback=IUPAC2Mol_back, coltitle=black, title=\textbf{IUPAC2Mol: Positional Isomer Disambiguation}]
\textbf{Input (IUPAC):} 2,6-dimethylbenzoic acid \\
\textbf{Ground Truth:} \texttt{CC1=C(C(=CC=C1)C)C(=O)O}

\textbf{Context Example:}
\begin{itemize}[leftmargin=1em]
    \item \textbf{IUPAC:} 2,4-dimethylbenzoic acid
    \item \textbf{SMILES:} \texttt{CC1=CC(=C(C=C1)C(=O)O)C}
\end{itemize}

\textbf{Comparison:}
\begin{itemize}[leftmargin=1em]
    \item \textbf{Direct:} \texttt{CC1=CC(=C(C=C1)C(=O)O)C} \quad (\textit{2,4-dimethyl isomer})
    \item \textbf{ICT:} \texttt{CC1=C(C(=CC=C1)C)C(=O)O} \quad \cmark
\end{itemize}
\end{tcolorbox}

\begin{tcolorbox}[colframe=IUPAC2Mol_front, colback=IUPAC2Mol_back, coltitle=black, title=\textbf{IUPAC2Mol: Carbon Chain Length Counting}]
\textbf{Input (IUPAC):} (9S,10S)-9,10-dihydroxyoctadecanoic acid \\
\textbf{Ground Truth:} \texttt{CCCCCCCC[C@@H]([C@H](CCCCCCCC(=O)O)O)O}

\textbf{Context Example:}
\begin{itemize}[leftmargin=1em]
    \item \textbf{IUPAC:} 9,10-dihydroxyoctadecanoic acid
    \item \textbf{SMILES:} \texttt{CCCCCCCCC(C(CCCCCCCC(=O)O)O)O}
\end{itemize}

\textbf{Comparison:}
\begin{itemize}[leftmargin=1em]
    \item \textbf{Direct:} \texttt{CCCCCCCCC[C@@H]([C@H](CCCCCCCC(=O)O)O)O} \quad (\textit{C19, extra carbon})
    \item \textbf{ICT:} \texttt{CCCCCCCC[C@@H]([C@H](CCCCCCCC(=O)O)O)O} \quad \cmark
\end{itemize}
\end{tcolorbox}

\begin{tcolorbox}[colframe=IUPAC2Mol_front, colback=IUPAC2Mol_back, coltitle=black, title=\textbf{IUPAC2Mol: Unsaturation Pattern Recognition}]
\textbf{Input (IUPAC):} (2E,4E,6S)-6-methylocta-2,4-dienoic acid \\
\textbf{Ground Truth:} \texttt{CC[C@H](C)/C=C/C=C/C(=O)O}

\textbf{Context Example:}
\begin{itemize}[leftmargin=1em]
    \item \textbf{IUPAC:} (2E,4E)-hexa-2,4-dienoic acid
    \item \textbf{SMILES:} \texttt{C/C=C/C=C/C(=O)O}
\end{itemize}

\textbf{Comparison:}
\begin{itemize}[leftmargin=1em]
    \item \textbf{Direct:} \texttt{C/C=C(\textbackslash C)/C=C/C(=O)O} \quad (\textit{wrong branching, missing ethyl})
    \item \textbf{ICT:} \texttt{CC[C@H](C)/C=C/C=C/C(=O)O} \quad \cmark
\end{itemize}
\end{tcolorbox}

\begin{tcolorbox}[colframe=IUPAC2Mol_front, colback=IUPAC2Mol_back, coltitle=black, title=\textbf{IUPAC2Mol: Heterocyclic Ring Connectivity}]
\textbf{Input (IUPAC):} 2-N-tert-butyl-6-chloro-1,3,5-triazine-2,4-diamine \\
\textbf{Ground Truth:} \texttt{CC(C)(C)NC1=NC(=NC(=N1)N)Cl}

\textbf{Context Example:}
\begin{itemize}[leftmargin=1em]
    \item \textbf{IUPAC:} 2-N-tert-butyl-6-chloro-4-N-ethyl-1,3,5-triazine-2,4-diamine
    \item \textbf{SMILES:} \texttt{CCNC1=NC(=NC(=N1)Cl)NC(C)(C)C}
\end{itemize}

\textbf{Comparison:}
\begin{itemize}[leftmargin=1em]
    \item \textbf{Direct:} \texttt{CC(C)(C)N1C(=NC(=N1)N)Cl} \quad (\textit{N bonded into ring})
    \item \textbf{ICT:} \texttt{CC(C)(C)NC1=NC(=NC(=N1)Cl)N} \quad \cmark
\end{itemize}
\end{tcolorbox}

\begin{tcolorbox}[colframe=IUPAC2Mol_front, colback=IUPAC2Mol_back, coltitle=black, title=\textbf{IUPAC2Mol: N-Oxide Functional Group}]
\textbf{Input (IUPAC):} 1-oxidopyridin-1-ium \\
\textbf{Ground Truth:} \texttt{C1=CC=[N+](C=C1)[O-]}

\textbf{Context Example:}
\begin{itemize}[leftmargin=1em]
    \item \textbf{IUPAC:} (5S)-1-methyl-5-(1-oxidopyridin-1-ium-3-yl)pyrrolidin-2-one
    \item \textbf{SMILES:} \texttt{CN1[C@@H](CCC1=O)C2=C[N+](=CC=C2)[O-]}
\end{itemize}

\textbf{Comparison:}
\begin{itemize}[leftmargin=1em]
    \item \textbf{Direct:} \texttt{C1=CC(=[N+](=C1)[O-])C} \quad (\textit{extra methyl, invalid SMILES})
    \item \textbf{ICT:} \texttt{C1=CC=[N+](C=C1)[O-]} \quad \cmark
\end{itemize}
\end{tcolorbox}

\begin{tcolorbox}[colframe=Mol2IUPAC_front, colback=Mol2IUPAC_back, coltitle=black, title=\textbf{Mol2IUPAC: Inorganic Salt Nomenclature}]
\textbf{Input (SMILES):} \texttt{[C-]\#N.[Ag+]} \\
\textbf{Ground Truth:} silver;cyanide

\textbf{Context Example:}
\begin{itemize}[leftmargin=1em]
    \item \textbf{SMILES:} \texttt{[C-]\#N.[Cu+]}
    \item \textbf{IUPAC:} copper(1+);cyanide
\end{itemize}

\textbf{Comparison:}
\begin{itemize}[leftmargin=1em]
    \item \textbf{Direct:} silver;cyanamide \quad (\textit{wrong anion name})
    \item \textbf{ICT:} silver;cyanide \quad \cmark
\end{itemize}
\end{tcolorbox}

\begin{tcolorbox}[colframe=Mol2IUPAC_front, colback=Mol2IUPAC_back, coltitle=black, title=\textbf{Mol2IUPAC: Substituent Position Identification}]
\textbf{Input (SMILES):} \texttt{CC1=C(C(=CC=C1)C)C(=O)O} \\
\textbf{Ground Truth:} 2,6-dimethylbenzoic acid

\textbf{Context Example:}
\begin{itemize}[leftmargin=1em]
    \item \textbf{SMILES:} \texttt{C1=CC(=C(C(=C1)O)C(=O)O)O}
    \item \textbf{IUPAC:} 2,6-dihydroxybenzoic acid
\end{itemize}

\textbf{Comparison:}
\begin{itemize}[leftmargin=1em]
    \item \textbf{Direct:} 2,3-dimethylbenzoic acid \quad (\textit{wrong positions})
    \item \textbf{ICT:} 2,6-dimethylbenzoic acid \quad \cmark
\end{itemize}
\end{tcolorbox}

\begin{tcolorbox}[colframe=Mol2IUPAC_front, colback=Mol2IUPAC_back, coltitle=black, title=\textbf{Mol2IUPAC: Ester vs.\ Nitro Group Naming}]
\textbf{Input (SMILES):} \texttt{CC(=O)OC1=CC=C(C=C1)[N+](=O)[O-]} \\
\textbf{Ground Truth:} (4-nitrophenyl) acetate

\textbf{Context Example:}
\begin{itemize}[leftmargin=1em]
    \item \textbf{SMILES:} \texttt{CCCC(=O)OC1=CC=C(C=C1)[N+](=O)[O-]}
    \item \textbf{IUPAC:} (4-nitrophenyl) butanoate
\end{itemize}

\textbf{Comparison:}
\begin{itemize}[leftmargin=1em]
    \item \textbf{Direct:} (4-acetyloxyphenyl) nitrate \quad (\textit{inverted parent/substituent})
    \item \textbf{ICT:} (4-nitrophenyl) acetate \quad \cmark
\end{itemize}
\end{tcolorbox}

\begin{tcolorbox}[colframe=Mol2IUPAC_front, colback=Mol2IUPAC_back, coltitle=black, title=\textbf{Mol2IUPAC: Acyl Chain Length Determination}]
\textbf{Input (SMILES):} \texttt{CCCCCCCCCCCCCCC(=O)OC(CC(=O)[O-])C[N+](C)(C)C} \\
\textbf{Ground Truth:} 3-pentadecanoyloxy-4-(trimethylazaniumyl)butanoate

\textbf{Context Example:}
\begin{itemize}[leftmargin=1em]
    \item \textbf{SMILES:} \texttt{CCCCCCCCCCCCCCCC(=O)O[C@H](CC(=O)[O-])C[N+](C)(C)C}
    \item \textbf{IUPAC:} (3R)-3-hexadecanoyloxy-4-(trimethylazaniumyl)butanoate
\end{itemize}

\textbf{Comparison:}
\begin{itemize}[leftmargin=1em]
    \item \textbf{Direct:} 3-\underline{hexadecanoyl}oxy-4-(trimethylazaniumyl)butanoate \quad (\textit{C16 instead of C15})
    \item \textbf{ICT:} 3-\underline{pentadecanoyl}oxy-4-(trimethylazaniumyl)butanoate \quad \cmark
\end{itemize}
\end{tcolorbox}

\begin{tcolorbox}[colframe=qed_front, colback=qed_back, coltitle=black, title=\textbf{QED Prediction: Halogenated Ester}]
\textbf{Input (SMILES):} \texttt{CC1([C@@H]([C@H]1C(=O)OCC2=C(C(=CC(=C2F)F)F)F)C=C(Cl)Cl)C} \\
\textbf{Ground Truth QED:} 0.4286

\textbf{Context Example:}
\begin{itemize}[leftmargin=1em]
    \item \textbf{SMILES:} \texttt{CC1=C(C(=C(C(=C1F)F)COC(=O)C2C(C2(C)C)/C=C(/C(F)(F)F)\textbackslash Cl)F)F}
    \item \textbf{QED:} 0.3638
\end{itemize}

\textbf{Comparison:}
\begin{itemize}[leftmargin=1em]
    \item \textbf{Direct:} 0.8127 \quad (error = 0.3841)
    \item \textbf{ICT:} 0.4417 \quad (error = 0.0131) \quad \cmark
\end{itemize}
\end{tcolorbox}

\begin{tcolorbox}[colframe=qed_front, colback=qed_back, coltitle=black, title=\textbf{QED Prediction: Prenylated Chalcone}]
\textbf{Input (SMILES):} \texttt{CC(=CCC1=C(C(=CC(=C1)/C=C/C(=O)C2=C(C=C(C=C2)O)O)OC)O)C} \\
\textbf{IUPAC:} (E)-1-(2,4-dihydroxyphenyl)-3-[4-hydroxy-3-methoxy-5-(3-methylbut-2-enyl)phenyl]prop-2-en-1-one \\
\textbf{Ground Truth QED:} 0.4107

\textbf{Context Example:}
\begin{itemize}[leftmargin=1em]
    \item \textbf{SMILES:} \texttt{CC(=CCC1=CC(=C(C=C1O)O)C(=O)/C=C/C2=CC=C(C=C2)O)C}
    \item \textbf{QED:} 0.4373
\end{itemize}

\textbf{Comparison:}
\begin{itemize}[leftmargin=1em]
    \item \textbf{Direct:} 0.6776 \quad (error = 0.2669)
    \item \textbf{ICT:} 0.4021 \quad (error = 0.0086) \quad \cmark
\end{itemize}
\end{tcolorbox}

\begin{tcolorbox}[colframe=qed_front, colback=qed_back, coltitle=black, title=\textbf{QED Prediction: Functional Group Variant}]
\textbf{Input (SMILES):} \texttt{C1=CC=C(C=C1)C(=O)C2=CC=CC(=C2N)CC(=O)N} \\
\textbf{IUPAC:} 2-(2-amino-3-benzoylphenyl)acetamide \\
\textbf{Ground Truth QED:} 0.6404

\textbf{Context Example:}
\begin{itemize}[leftmargin=1em]
    \item \textbf{SMILES:} \texttt{C1=CC=C(C=C1)C(=O)C2=CC=CC(=C2N)CC(=O)[O-]}
    \item \textbf{QED:} 0.6456
\end{itemize}

\textbf{Comparison:}
\begin{itemize}[leftmargin=1em]
    \item \textbf{Direct:} 0.8488 \quad (error = 0.2084)
    \item \textbf{ICT:} 0.6314 \quad (error = 0.0090) \quad \cmark
\end{itemize}
\end{tcolorbox}

\section{Prompt Templates}
\label{sec:prompt_templates}

We adopt two prompt formats depending on the training strategy.

\paragraph{Direct format.}
All six tasks share a unified conversational template:

\begin{tcolorbox}[colback=gray!5, colframe=gray!50, fontupper=\small\ttfamily, left=4pt, right=4pt, top=2pt, bottom=2pt]
\#\# User: \{task instruction\}: \{input\} \\
\#\# Assistant: \{target\}
\end{tcolorbox}

\noindent Table~\ref{tab:prompt_templates} lists the task-specific instructions.

\begin{table}[htbp]
    \centering
    \small
    \begin{tabular}{lll}
        \toprule
        \rowcolor{headergray}
        \textbf{Task} & \textbf{Instruction} & \textbf{Target} \\
        \midrule
        Mol2IUPAC & Generate the IUPAC name for the molecule: \texttt{\{SMILES\}} & IUPAC name \\
        IUPAC2Mol & Generate a molecule based on the IUPAC name of the molecule: \texttt{\{IUPAC\}} & SMILES \\
        QED & Predict the qed value for the molecule: \texttt{\{SMILES\}} & QED value \\
        LogP & Predict the logp value for the molecule: \texttt{\{SMILES\}} & LogP value \\
        HOMO-LUMO & Predict the HOMO-LUMO gap value for the molecule: \texttt{\{SMILES\}} & Gap value \\
        Polarizability & Predict the polarizability value for the molecule: \texttt{\{SMILES\}} & Polar.\ value \\
        \bottomrule
    \end{tabular}
     \caption{Task-specific prompt instructions and input/output types.}
     \label{tab:prompt_templates}
\end{table}

\paragraph{ICT format.}
When a sufficiently similar training example is retrieved, we prepend it as an in-context demonstration:

\begin{tcolorbox}[colback=gray!5, colframe=gray!50, fontupper=\small\ttfamily, left=4pt, right=4pt, top=2pt, bottom=2pt]
Example 1: \{neighbor\_input\} label \{neighbor\_output\} \\
Now, given \{query\_input\}, please generate the label:
\end{tcolorbox}

\noindent Here \texttt{neighbor\_input} and \texttt{neighbor\_output} are the input--output pair of the retrieved nearest neighbor, and \texttt{query\_input} is the test molecule's input representation (SMILES for Mol2IUPAC and property tasks; IUPAC name for IUPAC2Mol).
The target is the same as in the direct format.

\section{Limitations}
Beyond the limitations already discussed in the paper, including the dependence of ICT on retrieval quality and the weakening of context anchors at high GED, we explicitly acknowledge the following additional limitations.

\noindent\textbf{Representation scope.}
Our evaluation is limited to sequence-based molecular representations (SMILES and SELFIES). Graph-based molecular models, which operate directly on molecular topology without linearization, are not covered by our analysis and may exhibit qualitatively different robustness profiles under structural perturbation.

\noindent\textbf{Computational overhead of ICT.}
ICT requires a nearest-neighbor retrieval step at both training and inference time, adding computational overhead relative to standard fine-tuning. This may limit scalability to very large training sets or latency-sensitive deployment scenarios.

\section{Broader Impacts}
This work studies the manifold regularity of molecular LLMs under structural perturbations and proposes ICT as a manifold-regularization objective. On the positive side, our perturbation framework offers a practical diagnostic for the narrow ``trust region'' of current molecular LLMs, helping practitioners avoid over-relying on apparent in-distribution performance, which is particularly valuable for downstream applications such as drug discovery and materials design where over-confident predictions could lead to wasted experimental effort or safety concerns; ICT further provides a lightweight, retrieval-anchored route to improving robustness without architectural changes. On the negative side, more accurate molecular predictors carry a residual dual-use risk of being misused to design harmful compounds, though we view this risk as limited because the studied tasks (QED, LogP, HOMO–LUMO gap, polarizability, and IUPAC$\leftrightarrow$SMILES translation) are standard cheminformatics operations already supported by tools like RDKit, and ICT requires labeled training data in the target property domain. We recommend that any deployment be accompanied by domain expert oversight and explicit communication of the rapidly growing prediction uncertainty as inputs drift from the training manifold, an issue our work itself highlights.


\end{document}